\documentstyle[11pt]{article}

\renewcommand{\emptyset}{\O}
\newtheorem{theorem}{\sc Theorem}
\newtheorem{lemma}{\sc Lemma}
\newtheorem{coro}{\sc Corollary}
\newtheorem{nota}{\sc Notation}
\newtheorem{rem}{\sc Remark}
\newtheorem{defin}{\sc Definition}
\newtheorem{cla}{\sc Claim}
\newtheorem{ex}{\sc Example}
\newenvironment{proof}{\par \sc Proof.\rm}{\hspace*{\fill}$\Box$\vspace{1ex}}
\newenvironment{example}{\begin{ex}}{\hspace*{\fill}$\Diamond$\end{ex}}
\newenvironment{claim}{\begin{cla}}{\end{cla}}
\newenvironment{corollary}{\begin{coro}}{\end{coro}}
\newenvironment{definition}{\begin{defin}}{\end{defin}}

\newenvironment{remark}{\begin{rem}}{\hspace*{\fill}$\Diamond$\end{rem}}
\newenvironment{comment}{\begin{small}\begin{quotation}\hspace{-0.23in}\rm}{\end
{quotation}\end{small}}

\newcommand{\lea}{\stackrel{+}{<}}
\newcommand{\gea}{\stackrel{+}{>}}
\newcommand{\eqa}{\stackrel{+}{=}}

\title{Minimum Description Length Induction,  Bayesianism,
and Kolmogorov Complexity\thanks{Parts of the ideas in this paper
were announced in:
M. Li and P.M.B. Vitanyi, 
Computational Machine Learning in Theory and Praxis. In:
   `Computer Science Today', J. van Leeuwen, Ed., 
Lecture Notes in Computer Science, Vol.
   1000, Springer-Verlag, Heidelberg, 1995, 518-535; 
P.M.B. Vitanyi and M. Li, Ideal MDL and Its Relation To 
Bayesianism, `Proc. ISIS:
   Information, Statistics and Induction in Science', World Scientific,
Singapore, 1996, 282-291.}
}
\author{
Paul Vit\'{a}nyi\thanks{Partially supported by the European Union
through NeuroCOLT ESPRIT Working Group Nr. 8556,
by  NWO through NFI Project ALADDIN under Contract
number NF 62-376, and NSERC under
International Scientific Exchange Award ISE0125663.
Address: CWI,
Kruislaan 413, 1098 SJ Amsterdam, The Netherlands. Email: paulv@cwi.nl}\\
CWI and University of Amsterdam\\
\and
Ming Li\thanks{Supported in part by
the NSERC Operating Grant OGP0046506, ITRC, a CGAT grant, and the
Steacie Fellowship. 
Address: Department of Computer Science, University of Waterloo,
Waterloo, Ont. N2L 3G1, Canada. E-mail: mli@math.uwaterloo.ca}\\
University of Waterloo\\
}

\begin{document}
\maketitle
\begin{abstract}
The relationship between the Bayesian approach
and the minimum description length approach is established.
We sharpen and clarify the general modeling principles
MDL and MML, abstracted as the ideal MDL principle and defined from Bayes's
rule by means of Kolmogorov complexity. The basic condition
under which the ideal principle should be applied is encapsulated
as the Fundamental Inequality, which in broad terms states that the
principle is valid when the data are random, relative to every
contemplated hypothesis and also these hypotheses are random relative to
the (universal) prior. Basically, the ideal principle states
that the prior probability
associated with the hypothesis should be given by the algorithmic universal
probability, and the sum of the log universal probability
of the model plus the log of the probability of the data given the model
should be minimized. If we restrict the model class to the finite sets
then application of the ideal
principle 
turns into Kolmogorov's minimal sufficient statistic.
In general we show that data compression is almost always
the best strategy, both in hypothesis identification and prediction.

AMS Subject Classification:
Primary 68S05, 68T05; Secondary 62C10, 62A99.

Keywords: MDL,MML, Bayes's rule, Kolmogorov complexity, universal
distribution, randomness test

\end{abstract}

\section{Introduction}\label{sect.intro}

It is widely believed that the better a theory compresses the data
concerning some phenomenon under investigation, the better we have
learned, generalized, and the better the theory predicts unknown data.
This belief is vindicated in practice and is a form of
``Occam's razor'' paradigm about ``simplicity'' 
but apparently has not been rigorously
proved in a general setting. 
Here we show that
data compression is
almost always the best strategy, both
in hypotheses identification by using an ideal form 
of the minimum description length (MDL)
principle and in prediction of sequences.
To demonstrate these benificial aspects of compression
we use the Kolmogorov theory of complexity \cite{Ko65}
to express the optimal effective compression.
We identify precisely the situations in which MDL and Bayesianism
coincide and where they differ.

{\bf Hypothesis identification}
To demonstrate that compression is good for hypothesis identification  we
use the ideal MDL principle defined from Bayes's rule
by means of Kolmogorov complexity, Section~\ref{sect.ideal}.
This transformation is valid only for
individually random objects in computable distributions; 
if the contemplated objects are nonrandom or the distributions
are not computable then MDL and
Bayes's rule may part company. Basing MDL on first principles
we probe below the customary presentation of
MDL as being justified in and of itself
by philosophical persuasion  \cite{Ri78,Ri86}.  
The minimum message length (MML) approach,
while relying on priors, in practice is a related approach
\cite{Wa68,Wa87}.
Such approaches balance the complexity of the model
(and its tendency for overfitting) against the preciseness
of fitting the data (the error of the hypothesis).
Our analysis gives evidence why in practice Bayesianism is prone to
overfitting and MDL isn't.

{\bf Ideal MDL}
We are only interested in the following common idea shared between all
MDL-like methods: 
``Select the hypothesis which minimizes the sum of the
length of the description of the hypothesis (also called ``model'')
and the length of the description of the data relative to the
hypothesis.'' 
We take this to mean that every contemplated
individual hypothesis and every contemplated
 individual data sample is to be maximally
compressed: 
the description lengths involved should be the shortest
effective description lengths. We use ``effective'' in the sense
of ``Turing computable,'' \cite{Tu36}.
Shortest effective description length is asymptotically 
unique and objective and
known as the {\em Kolmogorov
complexity} \cite{Ko65} of the object being described. 
Thus, ``ideal MDL'' is a Kolmogorov complexity based form of the minimum
description length principle. 
In order to define ideal MDL from Bayes's rule we require some 
deep results due to L.A. Levin
\cite{Le74} and P. G\'acs \cite{Ga83} based on
the novel notion of individual randomness of objects
as expressed by P. Martin-L\"of's randomness tests \cite{ML66}. 
We show that the principle is valid
when a basic condition encapsulated as the ``Fundamental Inequality''
(\ref{eq.3}) in Section~\ref{sect.ideal} 
is satisfied. Broadly speaking this happens when the
data are random, relative to each contemplated hypothesis,
and also these hypotheses are random relative to the
contemplated prior. The latter requirement is always satisfied
for the so-called ``universal'' prior.
Under those conditions
ideal MDL, Bayesianism, MDL, and MML, select pretty
much the same hypothesis. 
Theorem~\ref{theo.mdlbayes} states that
minimum description length reasoning using shortest effective
descriptions coincides with Bayesian reasoning
using the universal prior distribution \cite{Le74,Ga74,Ch75}, 
provided the minimum
description length is achieved for those hypotheses with respect
to which the data sample is individually random (in the sense of Martin-L\"of).
If we restrict the model class to finite sets then this procedure
specializes to Kolmogorov's minimal sufficient statistics,
\cite{CT91,LiVibook}.

{\bf Kolmogorov complexity}
We recapitulate the basic definitions in Appendix~\ref{app.A}
in order to establish notation.
Shortest effective descriptions are ``effective''
in the sense that we can compute the described objects from them.
Unfortunately, \cite{Ko65,ZvLe70}, 
there is no general method to compute the length of a shortest description 
(the Kolmogorov complexity) from the object
being described. This obviously impedes actual use. Instead, one
needs to consider recursive approximations to shortest descriptions, 
for example
by restricting the allowable approximation time. This course is
followed in one sense or another in the practical incarnations such as
MML and MDL. There one often uses simply the Shannon-Fano code,
which assigns prefix code length $l_x := - \log P(x)$ to $x$ irrespective
of the regularities in $x$. If $P(x)=2^{-l_x}$ for
every $x \in \{0,1\}^n$, then the code word length of an
all-zero $x$ equals the code word length of a truly
irregular $x$. While the Shannon-Fano code gives an expected
code word length close to the entropy, it does not distinguish
the regular elements of a probability ensemble from the random
ones.  

{\bf Universal probability distribution}
Just as the Kolmogorov complexity measures the shortest effective
description length of an object, the universal probability
measures the greatest effective probability. Both notions are
objective and absolute in the sense of being recursively invariant 
by Church's thesis, \cite{LiVibook}. We give definitions in
Appendix~\ref{app.B}. We use universal probability
as a universal prior in Bayes's rule to analyze ideal MDL. 

{\bf Martin-L\"of randomness}
The common meaning of a ``random object'' is an outcome
of a random source. Such outcomes have expected properties
but particular outcomes may or may not possess these expected
properties. In contrast, we use
the notion of randomness of individual objects. This elusive notion's long
history goes back to the initial attempts by von Mises,
\cite{Mi19}, to formulate
the principles of application of the calculus of probabilities to
real-world phenomena. 
Classical probability theory
cannot even express the notion of ``randomness of individual objects.''
Following almost half a century of unsuccessful attempts,
the theory of Kolmogorov complexity, \cite{Ko65}, and Martin-L\"of tests 
for randomness, \cite{ML66}, finally succeeded in formally
expressing the novel notion of 
individual randomness in a correct manner, see \cite{LiVibook}. 
 Every individually random object 
possesses individually all effectively
testable properties that are only expected for outcomes of
the random source concerned. It will
satisfy {\em all} effective tests for randomness---
known and unknown alike. In Appendix~\ref{app.C} we recapitulate
the basics.

{\bf Two-part codes}
The prefix-code of the shortest effective descriptions
gives an expected code word length close to the entropy
and also compresses the regular objects until all regularity is
squeezed out. All shortest effective descriptions are
completely random themselves, without any regularity whatsoever.
The MDL idea of a two-part code for a body of data $D$
is natural from the perspective of Kolmogorov
complexity. 
If $D$ does not contain any regularities at all, then it consists
of purely random data and the hypothesis is precisely that.
Assume that the body of data $D$ contains regularities.
With help of a description of those regularities (a model) we can 
describe the data compactly. Assuming that the regularities can be represented
in an effective manner (that is, by a Turing machine),
we encode the data as a program for that machine. Squeezing
all effective regularity out of the data, we end up
with a Turing machine representing the meaningful regular
information in the data together with a program for
that Turing machine representing the remaining
meaningless randomness of the data. This intuition
finds its basis in the Definitions~\ref{def.KolmC}
and \ref{def.KolmK} in Appendix~\ref{app.A}. 
However, in general there are many ways
to make the division into meaningful information and remaining 
random information. In a painting
the represented image, the brush strokes, or even
finer detail can be the relevant information, 
depending on what we are interested in. What we require is
a rigorous mathematical condition to force a sensible division
of the information at hand in a meaningful part 
and a meaningless part. 
One way to do this in a restricted setting where the hypotheses 
are finite sets
was suggested by Kolmogorov at a Tallin conference in 1973
and published in \cite{Ko83}. 
See \cite{CT91,LiVibook}
and Section~\ref{app.KMSS}.
Given data $D$,
the goal is to identify
the ``most likely'' finite set $A$ of which $D$ is a ``typical'' element.
For this purpose we consider sets $A$ such that
$D \in A$ and we represent $A$ by the {\em shortest} program $A^*$ that
computes the characteristic function of $A$. The {\em Kolmogorov minimal
sufficient statistic} is the shortest $A^*$, say $A^*_0$ associated
with the set $A_0$, over all $A$ containing $D$
such that
the two-part description consisting of $A^*_0$ and $\log d(A_0)$ is as 
as short as the shortest {\em single} program that computes $D$ without
input. This definition is non-vacuous since
there is a two-part code (based on hypothesis $A_D = \{D\}$)
that is as concise as the shortest single code.

The shortest two-part code must be at least as long as the
shortest one-part code. Therefore, the description of $D$ given $A^*_0$
cannot be significantly shorter than $\log d(A_0)$. By
the theory of Martin-L\"of randomness in Appendix~\ref{app.C} this means
that $D$ is a ``typical'' element of $A$.
The ideal MDL principle expounded in this paper
is essentially a generalization of 
the Kolmogorov minimal sufficient statistic.

Note that in general 
finding a minimal sufficient statistic
is not recursive. Similarly, even
computing the MDL optimum in a much more restricted
class of models may run in computation
difficulties since it involves finding an optimum in a
large set of candidates.
In some cases
one can approximate this optimum, \cite{Vo95,Ya95}. 

{\bf Prediction}
The best single hypothesis does not
necessarily give the best prediction. For example, consider
a situation where we are given
a coin of unknown bias $p$ of coming up ``heads'' which is either
$p_1= \frac{1}{3}$ or $p_2=\frac{2}{3}$. Suppose we
have determined that there is probability $\frac{2}{3}$
that $p=p_1$ and probability $\frac{1}{3}$ that $p=p_2$.
Then the ``best'' hypothesis is the most likely one: $p=p_1$
which predicts a next outcome ``heads'' as having probability $\frac{1}{3}$.
Yet the best prediction is that this probability is the expectation
of throwing ``heads'' which is
\[ \frac{2}{3}p_1 + \frac{1}{3}p_2 = \frac{4}{9} . \]
 
Thus, the fact that compression is good for hypothesis identification
problems does not imply that compression is good for prediction.
In Section~\ref{sect.predict} we analyze the relation 
between compression of the data sample
and prediction in the very general
setting of R. Solomonoff \cite{So64,So78}.
We explain Solomonoff's prediction method using
the universal distribution. We show that this method is not
equivalent to the use of shortest descriptions. Nonetheless, we
demonstrate that compression of descriptions almost always gives optimal
prediction.

{\bf Scientific inference}
The philosopher D. Hume
(1711--1776) argued  \cite{Hu} that true induction
is impossible because we can only reach conclusions by using
known data and methods. Therefore, the conclusion is logically
already contained in the start configuration. Consequently,
the only form of induction possible
is deduction. Philosophers
have tried to find a way out of this deterministic conundrum by
appealing to probabilistic reasoning
such as using Bayes's rule \cite{TB}.
One problem with this is where the ``prior probability''
one uses has to come from.
Unsatisfactory solutions have been proposed by
philosophers like R. Carnap \cite{Ca50}
and K. Popper \cite{Po59}.
 
Essentially, combining the ideas of Epicurus,
Ockham, Bayes, and modern
computability theory,
Solomonoff \cite{So64,So78}
has successfully invented
a ``perfect'' theory of induction.
It incorporates Epicurus's multiple explanations idea, \cite{As84},
since no hypothesis that is still consistent with the data
will be eliminated. It incorporates Ockham's simplest explanation
idea since the hypotheses with low Kolmogorov
complexity are more probable.
The inductive reasoning is performed by means
of the mathematically sound rule of Bayes.

{\bf Comparison with Related Work}
Kolmogorov's minimal sufficient
statistics deals with hypothesis selection
where the considered hypotheses are finite sets of bounded cardinality.
Ideal MDL hypothesis selection
generalizes this procedure to arbitrary settings.
It is satisfying that our findings on ideal MDL confirm the 
validity of the ``real'' MDL principle
which rests on the idea of stochastic complexity. The latter
is defined in such a way that it represents the shortest
code length only for almost all data samples (stochastically speaking
the ``typical'' ones) for all models with real parameters in 
certain classes of probabilistic models except 
for a set of Lebesgue measure zero,
\cite{Ri86,Da95,MeFe95}. Similar results concerning
 probability density estimation by
MDL are given in \cite{BaCo91}.
These references consider probabilistic models and conditions.
We believe that in many current situations 
the models are inherently non-probabilistic
as, for  example, in the transmission of 
compressed images over noisy channels, \cite{SE97}.
Our algorithmic analysis of ideal MDL is about
such non-probabilistic
model settings as well as probabilistic ones
(provided they are recursive).
The results are derived in a 
nonprobabilistic manner entirely different
from the cited papers. 
It is remarkable that
there is a close agreement between the real properly
articulated MDL principle and our ideal one. 
The ideal MDL principle is valid in case the data
is individually random with respect to the contemplated hypothesis
and the latter is an individually random element of the contemplated prior.
Individually random objects are in a rigorous formal sense
``typical'' objects in a probability ensemble and together they
constitute allmost all such objects (all objects except
for a set of Lebesgue measure zero in the continuous case).
The nonprobabilistic expression of the range of validity of ``ideal MDL.'' 
implies the probabilistic expressions of the range
of validity of the ``real MDL'' principle.

Our results are more precise than the earlier probabilistic ones
in that they explicitly identify the ``excepted set
of Lebesgue measure zero'' 
 for which the principle may not be valid
as the set of ``individually nonrandom elements.''
The principle selects models such that the presented data are individually
random with respect to these models: if there is a true model and the
data are not random with respect to it then the principle avoids
this model.
This leads to a mathematical explanation
of correspondences and differences 
between ideal MDL and Bayesian reasoning,
and in particular it gives some
evidence under what conditions the latter is prone 
to overfitting while the former isn't.

\section{Ideal MDL}\label{sect.ideal}
The idea of predicting sequences using shortest
effective descriptions was first formulated by R. Solomonoff,
\cite{So64}. He uses Bayes's formula equipped with a fixed
``universal'' prior distribution.
In accordance with Occam's dictum,
that distribution gives most weight to the explanation that compresses the 
data the most. This approach inspired  Rissanen \cite{Ri78,Ri86} 
to formulate the MDL principle. Unaware of Solomonoff's work
Wallace and his co-authors \cite{Wa68,Wa87} formulated a related
but somewhat different
{\em Minimum Message Length (MML)} principle.

We focus only on the following central ideal version
which we believe is the essence of the matter. 
Indeed, we do not even care about whether
we deal with statistical or deterministic hypotheses.

\begin{definition}\label{def.mdl}
Given a sample of data, and an effective enumeration
of models, {\em ideal MDL} selects the model with the shortest
effective description that minimizes
the sum of
\begin{itemize}
\item
the length, in bits, of an effective description of the model; and
\item
the length, in bits, of an effective description of the data when encoded 
given the model. 
\end{itemize}
\end{definition}
\noindent
Under certain conditions on what constitutes a ``model'' 
and the notion of ``encoding given the model''
this coincides with
Kolmogorov's minimal sufficient statistic in Section~\ref{app.KMSS}.
In the latter a ``model'' 
is constrained to be a program that enumerates a finite set of data candidates
that includes the data at hand. 
Additionally, the ``effective description of the data
when encoded given the model'' is replaced by the logarithm
of the number of elements in the set that constitutes the model.
In ideal MDL we deal with model classes that may not be finite
sets, like the 
set of context-free languages or the set of recursive probability 
density functions. 
Therefore we require a more general approach.

Intuitively, a more complex 
hypothesis $H$ may fit the data better and therefore decreases the
misclassified data. If $H$ describes all the data,
then it does not allow for measuring errors.
A simpler description of
$H$ may be penalized by increasing the number of misclassified data.
If $H$ is a trivial hypothesis
that contains nothing, then
all data are described literally and there is no generalization.
The rationale of the method is that a balance in between
seems to be required. 

To derive the MDL approach we start from
{\em Bayes's rule} written as 
\begin{equation}\label{eq.Bayes}
\Pr(H |D)= {\Pr(D|H )P(H ) \over \Pr( D)}.
\end{equation}
\noindent
If the hypotheses space ${\cal H}$ is countable and the hypotheses $H$
are exhausive 
and mutually exclusive,
then
$\sum_{H \in {\cal H}} P(H) = 1$ 
$\Pr(D)= \sum_{H \in {\cal H}} \Pr(D|H )P(H )$.
For clarity and because it is relevant for the sequel
we distinguish notationally between the given
 {\em prior probability} ``$P( \cdot )$'' and 
the probabilities ``$\Pr ( \cdot )$'' that are induced by $P(\cdot)$
and the hypotheses $H$.
Bayes's rule maps input
($P(H),D$) to output $\Pr (H|D)$---the {\it posterior}
probability.
For many model classes (Bernoulli processes, Markov chains),
as the number $n$ of data generated by a true model in the class increases 
the total inferred probability can be expected to
concentrate on the `true' hypothesis (with probability one for
$n \rightarrow \infty$). That is, as $n$ grows the weight of 
the factor $\Pr(D|H)/\Pr (D)$ dominates the influence of the prior
$P(\cdot)$ for typical data---by
the law of large numbers. The importance of Bayes's rule 
is that the inferred probability
gives us as much information
as possible about the possible hypotheses
from only a small number of (typical) data
and the prior probability.

In general we
don't know the prior probabilities. The 
MDL approach in a sense replaces the unknown prior probability
that depends on the phenomenon being investigated 
by a fixed probability that depends on the coding used
to encode the hypotheses.
In ideal MDL the fixed ``universal'' probability (Appendix~\ref{app.B}) is
based on  Kolmogorov complexity---the length of the shortest
effective code (Appendix~\ref{app.A}).

In Bayes's rule we are concerned with
maximizing the term $\Pr(H|D)$ over $H$. Taking the
negative logarithm at both sides of the equation,
this is equivalent to {\it minimizing} the 
expression $ - \log  \Pr(H|D)$
 over $H$:
\begin{eqnarray*}
 - \log  \Pr(H|D) & = &- \log \Pr (D|H) - \log P(H) \\
                  &   &+ \log \Pr (D) .
\end{eqnarray*}
Since the probability $\Pr(D)$ is constant under varying $H$,
we want to find an $H_0$ such that
\begin{equation}\label{eq.bayes.mdl2}
H_0 := \mbox{\rm minarg}_{H \in {\cal H}} \{ - \log  \Pr(D|H) - \log  P(H) \}.
\end{equation}
In MML as in \cite{Wa87} or MDL as in \cite{Ri86} one roughly
interprets these negative logarithms of probabilities as the 
corresponding Shannon-Fano 
code word lengths. 
\footnote{
The term $- \log  \Pr(D|H)$ is also known as the %
{\it self-information}
\rm in information theory and the {\em negative 
log-likelihood} in statistics.
It can now be regarded as the number of bits it takes to redescribe
or encode $D$ with an ideal code relative to $H$. For the Shannon-Fano
code see Section~\ref{sect.ShFa}.}
But why 
not use 
the shortest effective descriptions with
code word length set equal to the Kolmogorov complexities?
This has an expected code word length about equal to the entropy,
\cite{LiVibook}, but additionally it compresses each object
by effectively squeezing out and accounting for all regularities in it.
The resulting code word is maximally random, that is, it
has maximal Kolmogorov complexity.
\footnote{
The relation between the Shannon-Fano code and Kolmogorov complexity
is treated in Section~\ref{sect.ShFa}.
For clarity of treatment, we refer the reader
to the Appendices or \cite{LiVibook} for all definitions and analysis of
auxiliary notions. This way we also do not deviate from the main argument,
do not obstruct the knowledgeable reader, and do
not confuse or discourage the reader who is unfamiliar
with Kolmogorov complexity theory. The bulk of the material is
Appendix~\ref{app.C} on Martin-L\"of's theory of randomness tests.
In particular the explicit expressions of universal 
randomness tests for arbitrary recursive distributions is
unpublished apart from \cite{LiVibook} and partially in \cite{Ga83}.}

Under certain constraints to be determined later,
the probabilities involved in (\ref{eq.bayes.mdl2}) can be substituted by
 the corresponding {\em universal probabilities}
$\bf m (\cdot)$ (Appendix~\ref{app.B}):
\begin{eqnarray}\label{eq.mdl}
\log P(H) & := & \log \hbox{\bf m} (H), \\
\nonumber
\log \Pr(D|H) & := & \log \hbox{\bf m} (D|H). 
\end{eqnarray}
According to \cite{Le74,Ga74,Ch75} we can then substitute
\begin{eqnarray}\label{eq.m=K}
- \log \hbox{\bf m} (H) & = & K(H), \\
\nonumber
- \log \hbox{\bf m} (D|H) & = & K(D|H) , 
\end{eqnarray}
where $K(\cdot)$ is the prefix complexity of Appendix~\ref{app.A}.
This way we replace the
sum of (\ref{eq.bayes.mdl2}) by the sum
of the minimum lengths
of effective
self-delimiting programs that compute
descriptions of $H$ and $D|H$.
The result is 
the code-independent, recursively invariant,
absolute form of the MDL principle:

\begin{definition}
\rm
Given an hypothesis class ${\cal H}$ and a data sample
$D$, the {\em ideal MDL} principle selects
the hypothesis
\begin{equation}\label{eq.minK}
H_0 :=  \mbox{minarg}_{H \in {\cal H}} 
\{ K(D|H)+K(H) \} .
\end{equation}
If there is more than one $H$ that minimizes (\ref{eq.minK})
then we break the tie by selecting the one of least complexity $K(H)$.
\end{definition}

The key question of Bayesianism versus ideal MDL is: When is
the substitution (\ref{eq.mdl}) valid?
We show that
in a simple setting were the hypotheses are finite sets
the ideal MDL principle and Bayesianism 
using the universal prior ${\bf m}(x)$
coincide with each other and with the Kolmogorov minimal
sufficient statistic. We generalize this to probabilistic hypothesis
classes. In full generality however, ideal MDL and Bayesianism
may diverge due to
the distinction between the $- \log P(\cdot)$
(the Shannon-Fano code length) and 
the Kolmogorov complexity $K(\cdot)$ (the shortest effective code length).
We establish the Fundamental Inequality defining the 
range of coincidence of the two principles.

  From now on, we will denote by $\lea$ an inequality to within an
additive constant, and by $\eqa$ the situation when both $\lea$ and
$\gea$ hold.

\subsection{Kolmogorov Minimal Sufficient Statistic}
\label{app.KMSS}
Considering only hypotheses that are finite sets of binary strings
of finite lengths the hypothesis selection principle known as 
``Kolmogorov's minimal sufficient statistic'' \cite{Ko83}
has a crisp formulation in terms
of Kolmogorov complexity. 
For this restricted hypothesis class we show that
the Kolmogorov minimal sufficient statistic 
is actually Bayesian hypothesis selection using
the universal distribution ${\bf m}(\cdot)$ as prior distribution and
it also coincides with the ideal MDL principle.

We follow the treatment of \cite{CT91,LiVibook} using
prefix complexity instead of plain complexity. 
Let $k$ and $\delta$ be natural numbers. A binary string $D$ representing
a data sample is called $( k , \delta )$-%
\it stochastic %
\index{string!$( k , \delta )$-stochastic}
\rm if there
is a finite set $H   \subseteq   \{0,1\}^*$ and $D \in H$ such that
$$
D \in H, \ \   K(H) \leq k , \ \  K(D|H) \geq \log d(H) - \delta .
$$
The first inequality (with $k$ not too large)
means that $H$ is sufficiently simple. The second
inequality (with the randomness deficiency $\delta$ not too large) means that
$D$ is an undistinguished (typical) element of $H$.
Indeed, if $D$ had properties defining
a very small subset $H'$ of $H$, then
these could be used to obtain a simple description of $D$
by determining its ordinal number in $H'$, which would
require $\log d(H')$ bits, which is much less than $\log d(H)$.
 
Suppose we carry out some probabilistic experiment
of which the outcome can be a priori every natural number.
Suppose this number is $D$. Knowing $D$, we want to
recover the probability distribution $P$ on 
the set of natural numbers ${\cal N}$. It seems reasonable to
require that first, $P$ has a simple description, 
and second, that $D$ would be a ``typical'' outcome of
an experiment with probability distribution $P$---that is, $D$ is
maximally random with respect to $P$.
The analysis above addresses a simplified form of this 
issue where $H$ plays the part of $P$---for example,
$H$ is a finite set of high-probability elements. In $n$ tosses
of a coin with probability $p>0$ of coming up ``heads,'' 
the set $H$ of outcomes consisting of 
binary strings of length $n$ with $n/2$ 1's constitutes a
set of cardinality ${n \choose {n/2}} = \Theta ( 2^n/\sqrt{n} )$.
To describe an element $D \in H$ requires $\lea n - \frac{1}{2} \log n$
bits. To describe $H \subseteq \{0,1\}^n$ given $n$ requires $O(1)$ bits 
(that is, $k$ is small in (\ref{eq.defKMSS}) below).
Conditioning everything on the length $n$, we have
$$ K(D|n) \lea K(D|H,n)+K(H|n)\lea n- \frac{1}{2} \log n , $$
and for the overwhelming majority of the $D$'s in $H$,
\[ K(D|n) \gea  n - \frac{1}{2} \log n . \]
These latter $D$'s are $(O(1),O(1))$-stochastic.

The {\em Kolmogorov structure function} $K_k (D|n)$ of
$D \in \{0,1\}^n$ is defined by 
\[ K_k (D|n) = \min \{ \log d(H): D \in H, \; \;   K(H|n) \leq k \} . \]
For a given small constant $c$, let 
$k_0$ be the least $k$ such that
\begin{equation}\label{eq.defKMSS}
 K_k(D|n)+k \leq K(D|n)+c . 
\end{equation}
Let $H_0$ be the corresponding set, and
let $H_0^*$ be its shortest program.
This $k_0$ with $K(H_0|n) \leq k_0$ is the least $k$ for which the two-part
description of $D$ is as parsimonious as the best single part
description of $D$. 

For this approach to be meaningful
we need to show that there always exists a $k$ satisfying (\ref{eq.defKMSS}).
For example, consider the hypothesis $H_D := \{D\}$.
Then, $\log d(H_D)=0$ and $K(H_D|n) \eqa K(D|n)$ which shows that
setting $k : \eqa K(D|n)$ satisfies (\ref{eq.defKMSS}) since $K_k (D|n) \eqa 0$.

If $D$ is maximally complex in some set $H_0$,
then $H_0$ represents all non-accidental structure in $D$. 
The $K_{k_0} (D|n)$-stage of the description just 
provides an index for $x$ in $H_0$---essentially the description of the
randomness or accidental structure of the string. 

Call a program $H^*$ a {\em sufficient statistics} if the complexity
$K(D|H,n)$ attains its maximum value $\eqa \log d(H)$ and
the randomness deficiency
$\delta (D|H) = \log d(H) - K(D|H,n)$ is minimal. 
\begin{definition}
\rm 
Let ${\cal H} := \{H: H \subseteq \{0,1\}^n \}$ and $D \in \{0,1\}^n$. Define
\begin{equation}\label{def.kmss}
 H_0 := \mbox{\rm minarg}_{H \in {\cal H}} \{K(H|n): K(H|n) + \log d(H) \eqa K(D|n) \}. 
\end{equation}
The set $H_0$---rather the shortest program $H_0^*$ that prints out
the characteristic sequence of $H_0 \in \{0,1\}^n$---is
called the {\em Kolmogorov minimal
sufficient statistic (KMSS)}\index{Kolmogorov minimal sufficient statistic|bold}
for $D$, given $n$. 
\end{definition}

All programs describing 
sets $H$ with $K(H|n) \leq k_0$ such that $K_{k_0}(D|n)+k_0 \eqa K(D|n)$
are sufficient statistics.
But the ``minimal'' 
sufficient statistic is induced by
the set $H_0$ having the shortest description among them.
Let us now tie the Kolmogorov minimal sufficient statistic to
Bayesian inference and ideal MDL. 
\begin{theorem}\label{theo.KMBIMDL}
Let $n$ be a large enough positive integer.
Consider the hypotheses class ${\cal H} := \{H: H \subseteq \{0,1\}^n \}$ and
a data sample $D \in \{0,1\}^n$.
All of the following principles select the same hypothesis:

{\rm (i)} Bayes's rule to select the least complexity hypothesis among
the hypotheses of maximal a posterior probability
using both (a) the universal distribution ${\bf m}(\cdot)$
as prior distribution, and (b) $\Pr(D|H)$ is the uniform probability 
$1/d(H)$ for $D \in H$ and 0 otherwise;

{\rm (ii)} 
Kolmogorov minimal sufficient statistic; and

{\rm (iii)} 
ideal MDL.
\end{theorem}

\begin{proof}
(i) $\leftrightarrow$ (ii). Substitute probabilities as in the statement
of the theorem in
(\ref{eq.bayes.mdl2}).

(ii) $\leftrightarrow$ (iii).
Let $H_0$ be the Kolmogorov minimal sufficient statistic for $D$
so that (\ref{def.kmss}) holds.
In addition, it is straightforward that $K(H_0|n)+K(D|H_0,n) \gea K(D|n)$,
and $K(D|H_0,n) \lea \log d(H_0)$ because we can describe $D$
by its index in the set $H_0$.
Altogether it follows that
$K(D|H_0,n) \eqa \log d(H_0)$ and
 \[
K(H_0|n)+ K(D|H_0,n) \eqa K(D|n).
\]
Since $K(H|n)+K(D|H,n) \gea K(D|n)$ for all $H \in {\cal H}$,
if \[
{\cal H}_0 = \{H': H' = \mbox{\rm minarg}_{H \in {\cal H}} \{K(H|n)+K(D|H,n)\} \} \]
then 
\[ 
H_0 = \mbox{\rm minarg}_H \{K(H): H \in {\cal H}_0 \}, 
\]
which is what we
had to prove.
\end{proof}

\begin{example}
\rm
Let us look at a coin toss example. If the probability $p$
of tossing ``1'' is unknown, then we can give a two-part description
of a string $D$ representing the sequence of $n$ outcomes by
describing the number $k$ of 1's in $D$ first, followed by the index 
$j \leq d(H)$  of $D$ in
in the set $H$ of strings with $k$ 1's. In this way ``$k|n$'' functions
as the model. If $k$ is incompressible
with $K(k|n) \eqa \log n$ and $K(j|k,n) \eqa \log {n \choose k}$
then the Kolmogorov minimal sufficient statistic
is described by $k|n$ in $\log n$ bits.
However if $p$ is a simple value like $\frac{1}{2}$ (or $1/\pi$), then
with overwhelming probability we obtain a much simpler
 Kolmogorov minimal sufficient
characteristic by a description of $p= \frac{1}{2}$ and 
$k = \frac{n}{2} + O( \sqrt{n})$  so that $K(k|n) \lea \frac{1}{2} \log n$.
\end{example}

\subsection{Probabilistic Generalization of KMSS}
Comparison of (\ref{eq.bayes.mdl2}), (\ref{def.kmss}),
 and Theorem~\ref{theo.KMBIMDL} suggests
a more general probabilistic version of Kolmogorov minimal sufficient
statistic.
This version turns out to coincide with maximum a posteriori Bayesian inference
but not necessarily with ideal MDL without additional conditions.
\begin{definition}\label{def.gkmss}
\rm
Let ${\cal H}$ be an enumerable class of probabilistic hypotheses 
and ${\cal D}$ be an enumerable domain of data samples such
that for every $H \in {\cal H}$ the probability density function
$\Pr ( \cdot |H)$ over the domain of data samples is recursive. 
Assume furthermore that for every data sample $D$ in the domain
there is an $H_D \in {\cal H}$ 
such that $\Pr (D|H_D) =1$ and $K(D|H_D) \eqa 0$ (the hypothesis
forces the data sample).
Define
\begin{equation}\label{eq.gkmss}
 H_0 := \mbox{\rm minarg}_{H \in {\cal H}} 
\{K(H): K(H) - \log \Pr (D|H) \eqa K(D) \}. 
\end{equation}
The set $H_0$---rather the shortest program $H_0^*$ that prints out
the characteristic sequence of $H_0 \in \{0,1\}^n$---is
called the {\em generalized Kolmogorov minimal
sufficient statistic (GKMSS)}
for $D$.
\end{definition}

The requirement that for every data sample in the domain there is
an hypothesis that forces it
ensures that $H_0$ as in definition~\ref{def.gkmss} exists.
\footnote{The equivalent hypothesis for a data sample $D$ in the setting of the 
Kolmogorov minimal sufficient statistic
was $H_D = \{D\}$.
}
\footnote{The Kolmogorov minimal sufficient statistic of Section~\ref{app.KMSS}
is the special case of the generalized
version for hypotheses that are finite sets and with
``$\Pr (D|H)$'' is the uniform probability  ``$1/d(H)$''.
}
\begin{theorem}\label{theo.gkmss}
The least complexity maximum a posteriori probability
hypothesis $H_0$ in Bayes's rule using prior
 $P(H) := {\bf m}( H)$
coincides with the generalized Kolmogorov minimal sufficient statistic.
\end{theorem}

\begin{proof}
Substitute $P(x) := {\bf m}(x)$ in
(\ref{eq.bayes.mdl2}).
Using (\ref{eq.m=K}) the least complexity hypothesis
satisfying the optimization problem
is:
\begin{equation}\label{eq.bayes.mdl3}
H_0 := \mbox{\rm minarg}_{H'} \{H': H' := 
\mbox{\rm minarg}_{H \in {\cal H}}\{ K(H) - \log  \Pr(D|H) \} \}.
\end{equation}
By assumptions in definition~\ref{def.gkmss}
 there is an $H_D$ 
such that
$K(H_D) - \log \Pr (D|H_D) \eqa K(D)$. It remains to show
that $ K(H) - \log  \Pr(D|H) \gea K(D)$ for all $H,D$.

It is straightforward that $K(H)+K(D|H) \gea K(D)$.
For recursive $\Pr ( \cdot | \cdot)$ it holds
that $l_D \eqa - \log \Pr(D|H)$ is the code length of the
effective Shannon-Fano prefix code 
(see Section~\ref{sect.ShFa} or \cite{CT91,LiVibook})
to recover $D$ given $H$. Since the prefix complexity
is the length of the shortest effective prefix code we have
$- \log \Pr(D|H) \gea K(D|H)$.
\end{proof}



\subsection{Shannon-Fano Code, Shortest Programs, and Randomness}
\label{sect.ShFa}
There is a tight connection between
prefix codes, probabilities, and notions of optimal codes.
The Shannon-Fano prefix code \cite{CT91} for
an ensemble of source words
with probability density $q$
has code word length $l_q(x) := - \log q(x)$ (up to rounding)
for source word $x$.
This code 
satisfies
\[ H(q) \leq \sum_x q(x)l_q (x) \leq H(q)+1 \]  
where $H(q)$ is the entropy of $q$.
By the Noiseless  Coding Theorem
this is the least expected code word length
among all prefix codes.
Therefore, the hypothesis $H$ which minimizes 
(\ref{eq.bayes.mdl2}) written as
\[ l_{\Pr(\cdot|H)} (D) + l_P (H))  \]
minimizes the sum of two prefix codes that both have shortest
{\em expected} code-word lengths. 
This is more or less what MML \cite{Wa87} and MDL \cite{Ri86} do.

But there are many prefix codes that have expected code word length
{\em almost} equal to the entropy. Consider only the class of prefix codes that
can be decoded by Turing machines (other codes are not practical).
There is is an optimal code in that class
with code word length $K(x)$ for object $x$. ``Optimality'' means that
for every prefix code in the class there is a constant $c$ such that
for all $x$ the length of the code for $x$ is at least $K(x)-c$, 
see Appendix~\ref{app.A} or \cite{LiVibook}.

In ideal MDL we minimize the sum of
the effective description lengths of the {\em individual} elements
$H,D$ involved
as in (\ref{eq.minK}). 
This is validated by Bayes's rule provided 
(\ref{eq.mdl}) holds.
To satisfy one part of (\ref{eq.mdl})
we are free to make the new
{\em assumption} that the prior probability $P( \cdot )$ in Bayes's rule
(\ref{eq.Bayes}) is fixed as ${\bf m}( \cdot )$. 
However, with respect to the other part of (\ref{eq.mdl}) we {\em cannot
assume} that the probability $\Pr ( \cdot |H)$ equals
${\bf m}( \cdot |H)$. Namely, 
probability $\Pr ( \cdot |H)$
may be totally determined by the hypothesis $H$. Depending on $H$
therefore, $l_{\Pr(\cdot|H)} (D)$ may be {\em very}
different from $K(D|H)$. This holds especially for `simple'
data $D$ which have low probability under assumption of hypothesis $H$. 

\begin{example}
\rm
Suppose we flip a coin of unknown bias $n$ times.
Let hypothesis $H$
and data $D$ be defined by:
\begin{eqnarray*}
H&:=&[\mbox{ Probability  `head' is $\frac{1}{2}$}] \\
D&:=&\underbrace{hh \ldots h}_{\mbox{$n$ times `$h$'(ead)s}}
\end{eqnarray*}
Then we have $\Pr(D|H)= 1/2^n$ and 
\[l_{\Pr(\cdot|H)} (D) =
-\log \Pr(D|H) = n.\]  
\noindent
In contrast, \[K(D|H) \lea \log n + 2 \log \log n.\]
\end{example}
The question arises exactly when is $- \log P(x) \eqa K(x)$? 
This is answered by the theory of individual randomness. Let 
$P: \{0,1\}^* \rightarrow [0,1]$ 
be a recursive probability density function.
\footnote{
A real-valued function is {\em recursive} if there is a Turing machine
that for every argument and precision parameter $b$
computes the function value within precision $2^{-b}$ and halts.}
By Theorem~\ref{PR3} (Appendix~\ref{app.C}) 
an element $x$ is Martin-L\"of random iff 
the universal test $\log ({\bf m}(x)/P(x)) \leq 0$.
That is, $- \log {\bf m}(x) \geq - \log P(x)$. 
\footnote{This means that {\em every} $x$ is random with respect to
the universal distribution ${\bf m}(x)$ (substitute $P(x):={\bf m}(x)$ above).}

\subsection{The Fundamental Inequality}

Let us call an hypothesis class ${\cal H}$ {\em rich enough} if
it contains a trivial hypothesis $H_{\emptyset}$ satisfying
$K(H_{\emptyset}) \eqa 0$ and satisfies the assumptions
of definition~\ref{def.gkmss}.
In ideal MDL as applied to such a rich enough hypothesis
class ${\cal H}$ 
there are two boundary cases:
The trivial hypothesis $H_{\emptyset}$
with $K(H_{\emptyset}) \eqa 0$ always implies 
that $K(D|H_{\emptyset}) \eqa K(D)$ and
therefore
$K(H_{\emptyset})+K(D|H_{\emptyset}) \eqa K(D)$. 
The hypothesis $H_D$ of definition~\ref{def.gkmss}
also yields $K(H_D)+K(D|H_D) \eqa K(D)$.
Since always $K(H)+K(D|H) \gea K(D)$,
these hypotheses minimize the ideal MDL description.

But for trivial hypotheses 
only Kolmogorov random data are typical. In fact,
ideal MDL correctly selects the trivial hypothesis
for individually random data.
But in general ``meaningful'' data are ``nonrandom''
in the sense that  $K(D) \ll l(D)$. But then $D$ is typical
only for nontrivial hypotheses, and a trivial hypothesis selected
by ideal MDL is not one for which the data are typical. We need 
to identify the conditions under which
ideal MDL restricts itself to selection
among hypotheses for which the given data are 
typical---it performs as the generalized Kolmogorov
minimal sufficient statistic.

Note that hypotheses 
satisfying (\ref{eq.gkmss}) may not always exist if we don't 
require that every data sample in the domain is forced by some
hypothesis in the hypothesis space we consider,
as we did in definition~\ref{def.gkmss}. 


\begin{example}
\rm
We look at a situation where the three optimization principles
(\ref{eq.bayes.mdl2}), (\ref{eq.minK}), (\ref{eq.gkmss}) act
differently.
Consider the outcome of $n$ trials of a Bernoulli process
$(p, 1-p)$. There are two hypotheses ${\cal H} = \{H_0, H_1\}$ where
\begin{eqnarray*} 
&& H_0 = [p = \frac{1}{2}] \\
&& H_1 = [p \neq \frac{1}{2}] 
\end{eqnarray*}
 The prior $P$ is
$P(H_0)=\frac{1}{2}$ and $P(H_1)=\frac{1}{2}$.
Consider the data sample
$D = 0^n$ with $n$ Kolmogorov
random (also with respect to $H_0$ and $H_1$)
so that $\log n \leq K(D), K(D|H_0), K(D|H_1) 
\leq \log n + 2 \log \log n$. Now 
\begin{eqnarray*}
&& - \log P(H_0) - \log \Pr(D|H_0) \eqa n \\
&& - \log P(H_1) - \log \Pr(D|H_1) \eqa 0  .
\end{eqnarray*}
Therefore Bayesianism selects $H_1$ which is intuitively correct.
Both hypotheses have complexity $\eqa 0$. Therefore we can
substitute $- \log P(H) := K(H)$ to obtain
\begin{eqnarray*}
&& K(H_0) - \log \Pr(D|H_0) \eqa n \\
&& K(H_1) - \log \Pr(D|H_1) \eqa 0  .
\end{eqnarray*}
Now generalized Kolmogorov minimal statistic doesn't select any hypothesis
at all because the right-hand side is unequal $K(D)$.
Ideal MDL on the other hand has the ex equo choice
\begin{eqnarray*}
&& K(H_0) +K(D|H_0) = \log n + O( \log \log n) \\
&& K(H_1) + K(D|H_1) = \log n + O(\log \log n) ,
\end{eqnarray*}
which intuitively seems incorrect. So we need to identify the conditions
under which ideal MDL draws correct conclusions.
\end{example}

While we can set the prior  $P( \cdot) := {\bf m}( \cdot)$ in Bayes's rule
to obtain the generalized Kolmogorov minimal sufficient statistic
for a rich enough hypothesis class,
we cannot in general also set $- \log \Pr(D|H_0) := K(D|H_0)$ 
to obtain ideal MDL.
%
The theory dealing with randomness of individual
objects states the  conditions for $-\log \Pr(D|H)$
and $K(D|H)$ to be close:
\footnote{This follows from (\ref{(3.4)}).}
\begin{itemize}
\item
Data sample $D$ is ``typical'' for $H \in {\cal H}$
iff $- \log \Pr (D|H) \lea K(D|H)$; and
\item
Data sample $D$ is ``not typical'' for $H \in {\cal H}$
iff  $- \log \Pr (D|H) \gg K(D|H)$.
\end{itemize}
Below we need the following:
\begin{definition}
\rm
Let $P: {\cal N} \rightarrow [0,1]$ be a 
recursive probability density distribution.
Then, the prefix complexity $K(P)$ of $P$
is defined as the length of the shortest self-delimiting program for 
the reference universal prefix machine to simulate the Turing machine
computing the probability density function $P$:
it is the shortest effective self-delimiting description of $P$,
(Appendix~\ref{app.A}).
\end{definition}
The relation between
Bayes's rule and ideal MDL governed by the following:
\begin{theorem}[Fundamental Inequality]
Let $\Pr( \cdot  | \cdot )$ and $P( \cdot )$ 
be recursive probability density functions. 

{\rm (i)}
If $D$ is  $\Pr( \cdot |H)$-random  and $H$ is $P(\cdot)$-random then
\begin{eqnarray}\label{eq.3}
  K(D|H)&+ &K(H) - \alpha(P,H) \\
\nonumber
& \leq & - \log \Pr(D|H) - \log P(H) \\
\nonumber
& \leq & K(D|H) +  K(H) ,
\end{eqnarray}
with 
\[ 
 \alpha(P,H)  =  K(\Pr(\cdot |H)) + K(P) .
\]

{\rm (ii)} If
\[ - \log \Pr(D|H) - \log P(H)  \eqa
  K(D|H) +  K(H) , \]
then  $D$ is  $\Pr( \cdot |H)$-random  and $H$ is $P(\cdot)$-random.
\end{theorem}

\begin{proof}
(i)
\begin{claim}\label{lem.delimDH}
If $D$ is a Martin-L\"of random element of 
the distribution $\Pr ( \cdot |H)$ 
then 
\begin{equation}\label{eq.delimDH}
K(D|H) - K(\Pr(\cdot |H))  \leq  - \log \Pr(D|H) 
 \leq  K(D|H) .
\end{equation}
\end{claim}
\begin{proof}
We appeal to the following known facts, Appendix~\ref{app.C}.
Because $\Pr (\cdot |H)$ is recursive:
${\bf m}(D|H) \geq 2^{-K(\Pr(\cdot|H))} \Pr(D|H)$,
(\ref{(3.4)}).
Therefore,
\begin{equation}\label{eq.1}
 \log \frac{{\bf m}(D|H)}{\Pr(D|H)}  \geq 
- K( \Pr ( \cdot |H)) .
\end{equation}
Note that $K( \Pr ( \cdot |H)) \lea K(H)$ because
from $H$ we can compute $\Pr(\cdot|H)$ by 
assumption on $\Pr(\cdot | \cdot)$.
Secondly, if $D$ is a Martin-L\"of random element of
the distribution $\Pr( \cdot |H)$, then 
 by Theorem~\ref{PR3}:
\begin{equation}\label{eq.2}
\log \frac{{\bf m}(D|H)}{\Pr(D|H)} \leq 0 .
\end{equation}
For $D$'s that are Martin-L\"of random,
(\ref{eq.1}, \ref{eq.2}) mean by
(\ref{eq.mdl}) that (\ref{eq.delimDH}) holds.

\end{proof}

If we set the {\em a priori} probability $P(H)$
of hypothesis $H$ to the universal probability then we
obtain  directly $- \log P(H) = {\bf m}(x)$. However, we do not
need to make this assumption. For a recursive prior $P(\cdot)$,
we can analyze the situation
when $H$ is random with respect to
$P(\cdot)$. 
\begin{claim}
If $H$ is a Martin-L\"of random element of
$P( \cdot )$
then
\begin{equation}\label{eq.4}
 K(H) - K(P) \leq - \log P(H) \leq K(H) .
\end{equation}
\end{claim}
\begin{proof}
Analogous to the proof of (\ref{eq.delimDH}).
\end{proof}

\noindent
Together  (\ref{eq.delimDH}, \ref{eq.4})
yield the theorem, part (i).

(ii) Follows from (\ref{eq.4}), (\ref{lem.delimDH}).
\end{proof}

\begin{remark}
\rm
We would like the theorem to hold for the overwhelming majority
of (data,hypothesis)-pairs. It can be shown that
a fixed fraction of all objects of every length are random
according to the universal randomness test for
finite objects (Theorem~\ref{PR3}, Appendix~\ref{app.C}). 
In the sample space of infinite binary sequences
a related randomness test shows infinite binary sequences
are random with probability one 
(Lemma~\ref{lem.infrand}, Section~\ref{sect.predict}).
We would like to state that
almost all finite objects are random  for all recursive distributions.
Although this is the case for many probability distributions
(for example ${\bf m}(\cdot )$) we can ensure it for all probability
distributions by relaxing the randomness condition.

Clearly, for {\em finite} sequences randomness viewed as absence
of regularities is a matter of degree: it doesn't make sense
to say that $x$ is random and $x$ with the first bit flipped is
non-random. Relaxing the arbitrary dividing line we 
call finite $x$'s of length $n$ 
say {\em weakly random}
if $- \log {\bf m}(x) \geq - \log P(x) - \log n$. 
This is more in line with Martin-L\"of's original universal randomness
test for finite binary strings 
which are a little weaker than $\log ({\bf m}(x)/P(x)) \leq 0$
(Lemma~\ref{lem.spt}, Appendix~\ref{app.C}).
Now
\begin{eqnarray*}
\sum_x P(x)2^{\log\frac{{\bf m}(x)}{P(x)}}
& = & \sum_x {\bf m}(x) \\
& \leq & 1 - \epsilon ,
\end{eqnarray*}
for some constant $\epsilon > 0$.
The last inequality
(Lemma 4.3.2 in \cite{LiVibook})
is essentially
due to the halting problem.
By Markov's inequality (Appendix~\ref{app.C}) with overwhelming
$P$-probability ($> 1-1/n$) an $x \in
\{0,1\}^*$ is weakly random.

If we relax
our notion of individual randomness to say weak randomness
then at least a fraction of $ 1-1/n$ of all binary
strings of length $n$ is weakly random.
This may cause an increase of $\alpha (P,H)$ by an additive logarithmic term 
(logarithmic in the length of $P$ and $H$).
\end{remark}

\begin{remark}[Range of Validity  of FI]
\rm
Hypothesis $H$ is $P$-random means that $H$ is ``typical''
for the prior distribution $P( \cdot )$
in the sense that it must not belong to any effective minority
(sets on which a minority 
of $P$-probability is concentrated). 
That is, hypothesis $H$
does not have any effectively testable
properties
that distinguish it from a majority. 
In \cite{LiVibook}
it is shown that
this is the set of $H$'s such that
$K(H) \eqa -\log P(H)$. 
In case $P(H) = {\bf m}(H)$, that is, the prior distribution
equals the universal distribution, then for {\em all} $H$
we have $K(H) \eqa -\log P(H)$, that is, all hypotheses are random with
respect to the universal distribution.

For other prior distributions
some hypotheses are random, and some other hypotheses
are nonrandom.
Let the possible hypotheses correspond to the
binary strings of length $n$, and let $P_n$ be the uniform
distribution that assigns probability $P_n (H)=1/2^n$
to every hypothesis $H$. Let us assume that the hypotheses are
coded as binary strings of length $n$, so that $H \in \{0,1\}^n$. 
Then, $H :=00 \ldots 0$
has low complexity: $K(H|n) \lea \log n$.
However, $-\log P_n(H) = n$. Therefore, by (\ref{eq.4}), $H$ is not
$P_n $-random. If we obtain $H$ by $n$ flips of a fair coin
however, then with overwhelming probability we
will have that $K(H|n) \eqa n $ and therefore
$-\log P_n (H) \eqa K(H|n)$ and $H$ is $P_n $-random.

That data sample $D$ is $\Pr(\cdot|H)$-random means that the data are
random with respect to the probability distribution
$\Pr(\cdot|H)$ induced by the hypothesis $H$.
Therefore, we require that the sample data $D$ 
are `typical', that is, `randomly
distributed' with respect to $\Pr(\cdot|H)$. 

\end{remark}

\begin{remark}[Optimal Hypothesis Doesn't Satisify FI]
\rm
The only way to violate the Fundamental Inequality
is that either $D$ is not $\Pr(\cdot |H)$-random
and therefore $-\log \Pr(D|H) \gg K(D|H)$,
or that $H$ is not $P$-random and therefore
$-\log P(H) \gg K(H)$. We give an example of the first case.
\footnote{This follows from (\ref{(3.4)}).}

Consider the identification of a 
Bernoulli process $B_p =( p, 1-p )$ ($0<p<1$)
that generates a given data sample $D \in \{0,1\}^n$.
Let
$\Pr( D |B_p,n)$ denote the
distribution of the outcome $D$ of $n$ trials
of the process $B_p$.
If the data $D$ are ``atypical''
like $D=00 \ldots 0$  ($n$ failures) for
$p = \frac{1}{2}$ and $n$ large,
then it violates the $\Pr(\cdot|B_{1/2},n)$-randomness
test (\ref{eq.2}) by having $- \log \Pr( D| B_{1/2}) =n$
and $K( D| B_{1/2}) \lea \log n +
2 \log \log n$.

%

Let the true hypothesis is $B_0$. 
The data sample $D = 00 \ldots 0$ is Martin-L\"of random with respect to
$B_0$.
In fact, for every $p$ the data sample generated by
$B_p$ is with overwhelming likelihood Martin-L\"of random with respect
to $B_p$. For such data samples,
if furthermore the
prior probability $P(\cdot) := {\bf m}(\cdot)$, then
the Fundamental Inequality
holds. If in fact $B_{1/2}$ would have been the true
hypothesis and we have obtained the same data $D = 00 \ldots 0$
($n$ zeros) as before, then
the Fundamental Inequality is violated for this true hypothesis and with
an appropriate prior Bayes's rule selects $B_{1/2}$ while MDL selects $B_0$. 
\end{remark}

\subsection{Ideal MDL and Bayesianism}
The best model or hypothesis
to explain the data is one 
that is a ``typical'' element
of the prior distribution and the data are ``typical'' for
the contemplated hypothesis---as 
prescribed by Kolmogorov's minimum sufficient
statistics. 
Thus, it is reasonable to contemplate
only admissible hypotheses as defined below in selecting the best one.
\begin{definition}
Given data sample $D$ and prior probability $P$, 
we call a hypothesis $H$ {\em admissible}
if $H$ is $P$-random and
$D$ is $\Pr( \cdot |H)$-random
(which implies that the Fundamental Inequality (\ref{eq.3}) holds).
\end{definition}
\begin{theorem}\label{theo.mdlbayes}
Let the data sample be $D$ and let the corresponding  set of 
admissible hypotheses be ${\cal H}_D \subseteq {\cal H}$.
Then the maximum a posteriori probability
hypothesis $H_{\mbox{\scriptsize bayes}} \in {\cal H}_D$ 
in Bayes's rule and the hypothesis $H_{\mbox{\scriptsize mdl}} \in {\cal H}_D$
selected by ideal MDL
are roughly equal:
\begin{eqnarray}\label{eq.rough=}
2^{-\alpha (P,H)} & \leq &
\frac{\Pr(H_{\mbox{\scriptsize mdl}}|D)}{\Pr(H_{\mbox{\scriptsize bayes}}|D )}  \leq  1 \\
\nonumber
 \alpha (P,H) & \geq &
K(D|H_{\mbox{\scriptsize mdl}})+K(H_{\mbox{\scriptsize mdl}}) \\
\nonumber
& -& K(D|H_{\mbox{\scriptsize bayes}}) - K(H_{\mbox{\scriptsize bayes}}) \geq 0 .
\end{eqnarray}
\end{theorem}
\begin{proof}
If in the Fundamental Inequality
$\alpha(P,H)$ is small then this
means that both the prior distribution $P$ is simple,
and that the probability distribution $\Pr(\cdot|H)$ 
over the data samples
induced by hypothesis $H$ is simple.
In contrast, if $\alpha(P,H)$ is large, which means
that either of the mentioned distributions is not
simple, for example when $K(\Pr(\cdot|H))=K(H)$
for complex $H$, then there may be some discrepancy.
Namely, in Bayes's rule our purpose is to maximize
$\Pr(H|D)$, and the hypothesis $H$ that minimizes
$K(D|H) +  K(H)$ also maximizes $\Pr(H|D)$ up
to a $2^{-\alpha (P,H)}$ multiplicative factor.
Conversely, the $H$ that maximizes $\Pr(H|D)$
also minimizes $K(D|H) +  K(H)$ up to an additive
term $\alpha (P,H)$. That is, with
\begin{eqnarray}
\label{eq.hmdl}
\nonumber
&&H_{\mbox{\scriptsize mdl}} := \mbox{minarg}_{H'}
\{K(H'): H' :=  \mbox{minarg}_{H \in {\cal H}_D} 
\{ K(D|H) +  K(H)\} \} \\
&&H_{\mbox{\scriptsize bayes}} :=  \mbox{minarg}_{H'}
\{K(H'): H' := \mbox{maxarg}_{H \in {\cal H}_D} 
\{\Pr(H|D)\} \}
\end{eqnarray}
we obtain (\ref{eq.rough=}) from (\ref{eq.3}).
\end{proof}

As a consequence, if $\alpha(P,H)$ is small enough and Bayes's rule selects
an admissible hypothesis, and so does ideal MDL, then both criteria
are (approximately) optimized by both selected hypotheses.

In order to identify application of MDL with application
of Bayes's rule on some prior distribution $P$ as in Theorem~\ref{theo.mdlbayes}
we must assume that, given $D$, the Fundamental Inequality 
is satisfied for $H_{\mbox{\scriptsize mdl}}$ as defined in
(\ref{eq.hmdl}). This means that
$H_{\mbox{\scriptsize mdl}}$ is $P$-random for the used
prior distribution $P$. One choice to guarantee
this is 
\[ P(\cdot):= {\bf m}(\cdot) (= 2^{-K(\cdot)}).\]
This is a valid choice even though ${\bf m}$ is not
recursive. Namely, we only require that
${\bf m}(\cdot)/P(\cdot)$ 
be enumerable (Definition~\ref{def.enum.funct}, Appendix~\ref{app.B}), 
which is certainly guaranteed
by choice of $P(\cdot):={\bf m}(\cdot)$. The
choice of ${\bf m}(\cdot)$ as prior 
is an objective and recursively invariant 
Occam's razor: simple hypothesis $H$ (with
$K(H) \ll l(H)$ have high ${\bf m}$-probability, and
complex or random hypothesis H (with $K(H) \approx l(H)$)
have low ${\bf m}$-probability $2^{-l(H)}$.
The randomness test $\log ({\bf m}(H)/P(H))$ evaluates to $0$ for {\em every}
$H$, which means that all hypotheses
are random with respect to distribution ${\bf m}(\cdot)$.

\begin{theorem}\label{theo.IMDLKMSS}
Let $\alpha (P,H)$ in the FI (\ref{eq.3}) be small (for example $\alpha \eqa 0$)
and prior $P(\cdot) := {\bf m}(\cdot)$. Then
the Fundamental Inequality (\ref{eq.3}) is satisfied
iff data sample $D$ is (almost)
$\Pr(\cdot|H_{\mbox{\scriptsize mdl}})$-random.
\end{theorem}
\begin{proof}
With $\alpha (P,H) \eqa 0$ and $P(\cdot) := {\bf m}(\cdot)$
(so $- \log P(H) = K(H)$ by (\ref{eq.m=K}))
we can rewrite (\ref{eq.3}) as
\[ - \log \Pr (D|H) \eqa K(D|H) . \]
The theorem follows by Theorem~\ref{PR3} in Appendix~\ref{app.C}.
\end{proof}

If there is a true probabilistic model $H$ then with high probability
 the data sample $D$ will
be $\Pr ( \cdot |H)$-random.  This suggests that in selecting
the best model we should
contemplate only those models $H$ for which this is the case.
The requirement that $D$ be $\Pr(\cdot |H_{\mbox{\scriptsize mdl}})$-random
constrains the domain of hypotheses from which we can
choose $H_{\mbox{\scriptsize mdl}}$.  
\begin{corollary}
With high probability
ideal MDL is an application of Bayes's rule with 
the universal prior distribution
${\bf m}(\cdot)$ and selection of an optimal admissible
 hypothesis $H_{\mbox{\scriptsize mdl}}$
such that the data sample $D$ is
$\Pr(\cdot|H_{\mbox{\scriptsize mdl}})$-random (in the
sense of Appendix~\ref{app.C}).
\end{corollary}

Since the notion of individual randomness incorporates
all effectively testable properties of randomness (in the finite case
only to some degree),
application of ideal MDL will select 
the simplest hypothesis $H$
that balances $K(D|H)$ and $K(H)$ such that
the data sample $D$ is random to it---as far as can effectively be ascertained.

Restricted to the class of admissible hypotheses,
ideal MDL does not simply select the hypothesis that
precisely fits the data but it selects an hypothesis
that would typically generate the data. With
some amount of overstatement one can say that if one
obtains perfect data for a true hypothesis, then ideal MDL interprets
these data as data obtained from a simpler hypothesis
subject to measuring errors.
Consequently, in this case ideal MDL is going to 
give you the {\em false simple}
hypothesis and {\em not} the {\em complex true} hypothesis.

\begin{itemize}
\item
Ideal MDL only gives us the true hypothesis if the
data satisfies certain conditions relative to
the true hypothesis. Stated differently:
there are only data and no true hypothesis for ideal MDL.
The principle simply obtains the hypothesis suggested
by the data and it assumes that the data
are random with respect to the hypothesis.
\end{itemize}

\subsection{Applications}
Unfortunately, the function $K$ is not computable,
\cite{LiVibook}.
For practical applications 
one must settle for easily computable approximations,
for example, restricted model classes and particular codings
of hypotheses.
In this paper we will not address the question
which encoding one uses in practice, but refer
to references \cite{Ri86,Wa68,Ya95,Vo95}.

In statistical applications,
$H$ is some statistical distribution (or model)
$H = P ( \theta ) $ with a list of parameters
$\theta =  ( \theta_1 ,  \ldots  , \theta_k  ) $,
where the number $k$ may vary
and influence the (descriptional) complexity
of $ \theta $. (For example, $H$ can be
a normal distribution
\index{distribution!normal} $N( \mu , \sigma )$
described by $\theta = ( \mu , \sigma )$.)
Each parameter $\theta_i$ is truncated to
fixed finite precision.
The data sample consists of
$n$ outcomes ${\bf y}= (y_1 , \ldots , x_n)$
of $n$ trials  ${\bf x} = (x_1, \ldots , x_n)$
for distribution $P(\theta ) $. The
data $D$ in the above formulas is given as
$D=({\bf x}, {\bf y})$. By expansion of
conditional probabilities we have therefore
\[
\Pr (D|H) = \Pr ({\bf x}, {\bf y}|H)
= \Pr({\bf x} |H) \cdot \Pr ({\bf y}|H, {\bf x} ).
\]
In the argument above we take the negative logarithm
of $\Pr(D|H)$, that is,
\[ - \log \Pr (D|H) =  
- \log \Pr({\bf x} |H) - \log  \Pr ({\bf y}|H, {\bf x} ) .
\]
Taking the negative logarithm in Bayes's rule and
the analysis of the previous section now yields
that MDL selects the hypothesis with highest inferred
probability satisfying ${\bf x}$ is $\Pr(\cdot|H)$-random and 
${\bf y}$ is $\Pr(\cdot|H,{\bf x})$-random.
Bayesian reasoning selects the same hypothesis provided
the hypothesis with maximal inferred probability has
${\bf x},{\bf y}$ satisfy the same conditions.

\begin{example}[Learning Polynomials]
\rm
We wish to fit a
polynomial $f$ of unknown degree to a set of data points
$ D $
such that it can predict future data $y$ given $x$.
Even if the data did come from a polynomial curve of degree, say two,
because of measurement errors and noise,
we still cannot find a polynomial of degree two
fitting all $n$ points exactly.
In general, the higher the degree of fitting polynomial, 
the greater the precision of the fit. For $n$ data points,
a polynomial of degree $n-1$ can be made to fit exactly,
but probably has no predictive value.
Applying
ideal MDL we look for
$H_{\mbox{\scriptsize mdl}}   :=  
\mbox{minarg}_H \{ K({\bf x},{\bf y}|H) +  K(H)\}$.

Let us apply the ideal MDL principle where we describe all
$(k-1)$-degree polynomials by a vector of $k$
entries, each entry with a precision of $d$ bits.
Then, the entire polynomial is described by 
\begin{equation}\label{eq.costenc}
kd + O(\log kd) \mbox{  bits.}
\end{equation}
(We
have to describe $k$, $d$, and account for self-delimiting
encoding of the separate items.)
For example, $ax^2+bx+c$ is described by
$(a,b,c)$ and
can be encoded by about
$3d$ bits. Each datapoint $(x_i,y_i)$ which needs
to be encoded separately with precision of $d$ bits
per coordinate costs about $2d$ bits.

For simplicity assume that probability $\Pr ({\bf x}|H)=1$
(because ${\bf x}$ is prescribed).
To apply the ideal MDL principle we must trade
the cost of hypothesis $H$ (\ref{eq.costenc})
against the cost of describing ${\bf y}$ given $H$
and ${\bf x}$. As a trivial example, suppose 
$n-1$ out of $n$ datapoints fit a polynomial of degree 2 exactly,
but only 2 points lie on any polynomial of degree 1
(a straight line). Of course, there is a polynomial
of degree $n-1$ which fits the data precisely (up to
precision).  Then the ideal MDL cost is $3d + 2d$ for the
2nd degree polynomial, $2d + (n-2)d$ for the 1st degree
polynomial, and $nd$ for the $(n-1)$th degree polynomial.
Given the choice among those three options, 
we select the 2nd degree polynomial
for all  $n > 5$. 
\end{example}

\begin{remark}[Exception-Based MDL]
\rm
A hypothesis $H$ minimizing
$K(D|H) + K(H)$ always satisfies
\[ K(D|H) + K(H) \geq K(D).\]
Let $E \subseteq D$ denote the subset of the data 
that are {\em exceptions} to $H$ in the sense
of not being classified correctly by $H$.
The following {\em exception-based} MDL (E-MDL)
is sometimes confused with MDL:
With $E:=D-D_H$ and
$D_H$ is the data set classified
according to $H$, select 
\[ H_{\mbox{e-mdl}} =  \mbox{minarg}_{H'}
\{K(H'): H' :=  \mbox{minarg}_H \{ K(H)+ K(E|H) \} \} .\]
In E-MDL we look for the shortest description of an accepting program
for the data consisting of a classification rule $H$
and an exception list $E$.
While this principle sometimes
gives good results, application may lead to absurdity
as the following shows:

In many problems the data sample consists of 
positive examples only. For example, in learning (a grammar for)
English language, given 
the {\em Oxford English Dictionary}.
According
to E-MDL the best hypothesis is the
trivial grammar $H$ generating {\em all} sentences
over the alphabet. Namely, this grammar gives
$K(H) \eqa 0$ independent of $D$ and
also $E:= \emptyset$. Consequently,
\[ \min_H \{ K(H) + K(E|H) \}= K(H) \eqa 0 , \]
which is absurd. 
The E-MDL principle is vindicated and
reduces to standard MDL in the context
of interpreting $D=({\bf x},{\bf y})$
with ${\bf x}$ fixed as in ``supervised learning.''
Now
for constant $ K({\bf x}|H)$ 
 \[ H_{\mbox{e-mdl}} = 
 \mbox{minarg}_{H'}
\{K(H'): H' := \mbox{minarg}_H \{ K(H) + K({\bf y}|H,{\bf x}) 
+ K({\bf x}|H) \} \} \]
is the same as 
\[ H_{\mbox{mdl}} =   \mbox{minarg}_{H'}
\{K(H'): H' := \mbox{minarg}_H \{K(H) + K({\bf y}|H,{\bf x}) \} \}.\]
Ignoring the constant ${\bf x}$ in the conditional
$K({\bf y}|H,{\bf x})$ corresponds to $K(E|H)$.
\end{remark}


\section{Prediction by Minimum Description Length}
\label{sect.predict}
Let us consider theory formation
in science as the process of obtaining
a compact description of past observations together
with predictions of future ones.
Ray Solomonoff~\cite{So64,So78} argues 
that the preliminary data of the investigator,
the hypotheses he proposes,
the experimental setup he designs, the trials he performs,
the outcomes he obtains, the new hypotheses he formulates, and so on,
can all be encoded as the initial segment of a potentially
infinite binary sequence.
The investigator obtains increasingly
longer initial segments of an infinite binary
sequence $\omega$ by performing more and more
experiments on some aspect of nature.
To describe the underlying regularity of $\omega$, the investigator
tries to formulate a theory that governs $\omega$
on the basis of the
outcome of past experiments. Candidate theories (hypotheses)
are identified with computer programs that compute binary sequences
starting with the observed initial segment.

There are many different possible infinite sequences (histories)
on which the investigator can embark. The phenomenon he wants
to understand or the strategy he uses can be stochastic.
Each such sequence
corresponds to one never-ending sequential
history of conjectures and refutations and
confirmations and each initial segment has
different continuations governed by certain probabilities.
In this view each phenomenon can be identified with
a measure $\mu$ on the continuous sample
space of infinite sequences over a basic description alphabet.
This distribution $\mu$ can be said to be the concept
or phenomenon involved.
Now the aim is to {\em predict} outcomes concerning a phenomenon $\mu$
under investigation. In this case we have some prior evidence
(prior distribution over the hypotheses, experimental data)
and we want to predict future events. 

This situation can be
modelled by considering a sample space 
$S$ of one-way infinite sequences of basic elements ${\cal B}$
defined by $S={\cal B}^{\infty}$. We assume a
prior distribution\index{probability!prior} $\mu$ over
$S$
with $\mu (x)$ denoting
the probability of a sequence starting with $x$.
Here $\mu (\cdot )$ is a {\em semimeasure}\footnote{Traditional
notation is ``$\mu (\Gamma_x )$'' instead of ``$\mu (x)$'' 
where {\em cylinder}  $\Gamma_x = \{ \omega \in S:
\omega \mbox{ starts with } x \}$. We use ``$\mu (x)$'' for convenience.
$\mu$ is a {\em measure} if equalities hold.}
satisfying
\begin{eqnarray*}
\mu (\epsilon ) & \leq & 1 \\
\mu (x) & \geq & \sum_{a \in {\cal B}} \mu (xa).
\end{eqnarray*}
Given a previously observed data string $x$,
the inference problem is to predict the
next symbol in the output sequence,
that is, to extrapolate the sequence $x$.
In terms of the variables in
(\ref{eq.Bayes}), $H_{xy}$ is the hypothesis that
the sequence starts with initial segment $xy$.
Data $D_x$ consists of the fact that the sequence
starts with initial segment $x$.  Then,
$\Pr (D_x|H_{xy}) = 1$, that is, the data is forced by the hypothesis,
or $\Pr (D_z|H_{xy}) = 0$ for $z$ is not a prefix of $xy$, 
that is, the hypothesis contradicts the data. For
$ P (H_{xy} )$ and $\Pr(D_x)$ in (\ref{eq.Bayes}) we
substitute $\mu (xy)$ and $\mu (x)$, respectively.
For $\Pr(H_{xy} |D_x)$ we substitute $\mu (y|x)$.
This way   (\ref{eq.Bayes}) is rewritten as
\begin{equation}\label{eq3}
\mu (y|x)= { {\mu (xy)} \over {\mu (x)}} .
\end{equation}
The final probability
$\mu (y|x)$ is the probability of the next symbol string being $y$, given
the initial string $x$. Obviously we now only need the prior
probability $\mu$ to evaluate $\mu (y|x)$.
The goal of inductive inference in general is to
be able to either (i) predict, or extrapolate, the next
element after $x$ or (ii) to infer an underlying effective process
that generated $x$, and hence
to be able to predict the next symbol.
In the most general deterministic case such an effective
process is a Turing\index{machine!deterministic Turing}
machine,
but it can also be a probabilistic Turing machine
or, say,
a Markov process.
The central task of
inductive inference is to find a universally valid approximation
to $\mu$ which is good at
estimating the conditional probability that a
given segment $x$ will be followed by a segment $y$.

In general this is impossible.
But suppose we restrict the class of priors $\mu$ to the
{\em recursive} semimeasures\footnote{
there is a Turing machine
that for every $x$ and $b$ computes $\mu (x)$ within precision $2^{-b}$.}
and restrict the set of basic elements 
to $\{0,1\}$.
Under this relatively mild restriction on the admissible
semimeasures $\mu$, it turns out that
we can use the {\em universal semimeasure} ${\bf M}$
as a ``universal prior'' (replacing the real prior $\mu$) for prediction.
The theory of the  universal semimeasure
${\bf M}$, the analogue in the sample space $\{0,1\}^{\infty}$
of ${\bf m}$ in the sample space $\{0,1\}^*$ equivalent to ${\cal N}$,
 is developed in \cite{LiVibook}, Chapter 4, and Chapter 5.
It is defined with respect to a special
type Turing machine called {\em monotone} Turing machine.
The universal semimeasure ${\bf M}$ multiplicatively dominates all
enumerable (Definition~\ref{def.enum.funct}, Appendix~\ref{app.B}) 
semimeasures. It can be shown that if we flip
a fair coin to generate the successive bits on the input tape
of the universal reference monotone Turing machine, then 
the probability that it outputs $x \alpha$ ($x$
followed by something) is ${\bf M}(x)$, \cite{ZvLe70}.

The universal probability ${\bf M}( \cdot)$ allows us to
explicitly express
a universal randomness test for the elements in $\{0,1\}^{\infty}$
analogous to the universal randomness tests for the finite
elements of $\{0,1\}^*$ developed in Appendix~\ref{app.C}.
This notion of randomness
with respect to a recursive semimeasure $\mu$ satisfies the following
explicit characterization of a universal (sequential)
randomness test (for proof see \cite{LiVibook}, Chapter 4):
\begin{lemma}\label{lem.infrand}
Let $\mu$ be a recursive semimeasure on $\{0,1\}^{\infty}$.
An infinite binary sequence  $\omega$ is $\mu$-random iff
$$\sup_n {\bf M} (\omega_1 \ldots \omega_n)/\mu  (\omega_1 \ldots \omega_n)
< \infty ,$$
and the set of $\mu$-random sequences has $\mu$-measure one,
\end{lemma}
In contrast with the discrete case, the elements of $\{0,1\}^{\infty}$
can be sharply divided in the random ones that pass
{\em all} effective (sequential) randomness tests
 and the nonrandom ones that do not. 

We start by demonstrating
convergence of ${\bf M}(y|x)$
and $\mu (y|x)$ for $x \rightarrow \infty$, with $\mu$-probability 1.
\footnote{
We can express the ``goodness'' of predictions according
to ${\bf M}$ with respect to a true $\mu$ as follows:
Let $S_n$ be the $\mu$-expected value of the square
of the difference in $\mu$-probability and
${\bf M}$-probability of 0 occurring
at the $n$th prediction
$$
S_n = \sum_{l(x)=n-1} \mu (x)
({\bf M} (0 |x)  -  \mu (0 |x))^2 .
$$
We may call $S_n$ the
{\it expected squared error at the $n$th prediction}.
The following celebrated result of Solomonoff, \cite{So78}, says
that ${\bf M}$ is very suitable for prediction
(a proof using Kulback-Leibler divergence
is given in  \cite{LiVibook}):
\begin{theorem}
\label{PR9}
Let $\mu$ be a recursive semimeasure. Using the notation above,
$\sum_n  S_n    \leq  k/2$ with $k = K( \mu ) \ln 2$.
{\rm (}Hence, $S_n$ converges to 0 faster than $1/n$.{\rm )}
\end{theorem}
However, Solomonoff's result is not strong enough to give
the required
convergence of conditional probabilities with $\mu$-probability 1.
}
\begin{theorem}
\label{PR10}
Let $\mu$
be a positive recursive measure.
If the length of $y$ is fixed and the length of $x$
grows to infinity,
then
$$
{{\bf M} (y  | x )  \over \mu (y | x )}   \rightarrow   1,
$$
with $\mu$-probability one. The infinite sequences $\omega$
with prefixes $x$ satisfying the displayed asymptotics
are precisely the $\mu$-random sequences.
\end{theorem}
\begin{proof}
We use an approach
based on the Submartingale Convergence
Theorem, \cite{Do53} pp. 324-325,
which states that the following property holds for
each sequence of random variables
$\omega_1 , \omega_2 , \ldots $.
If $f ( \omega_{1:n} )$ is a $\mu$-submartingale, and
the $\mu$-expectation ${\bf E} |f( \omega_{1:n} )|   <   \infty$,
then it follows that
$\lim_{{n}   \rightarrow   \infty} f ( \omega_{1:n} )$ exists with
$\mu$-probability one.

In our case,
$$
t( \omega_{1:n} | \mu) = {{\bf M} ( \omega_{1:n} )  \over \mu ( \omega_{1:n} )}$$
is a $\mu$-submartingale, and the $\mu$-expectation
${\bf E}  t( \omega_{1:n}| \mu )   \leq  1$.
Therefore,  there is a set
$A   \subseteq   \{0,1\}^{\infty}$ with $\mu (A) = 1$, such that
for each $\omega \in A$
the limit
$\lim_{{n}   \rightarrow   \infty}  t( \omega_{1:n} | \mu) < \infty.$
These are the $\mu$-random $\omega$'s by Corollary 4.5.5 in
\cite{LiVibook}.
Consequently, for fixed $m$, for each $\omega$ in $A$, we have
$$
\lim_{{n}   \rightarrow   \infty} { {\bf M}
( \omega_{1:n+m} )/ \mu ( \omega_{1:n+m} )
\over  {\bf M} ( \omega_{1:n} )/ \mu ( \omega_{1:n} )} = 1,
$$
provided the limit of the denominator is not zero.
The latter fact is guarantied by the universality of ${\bf M}$:
for every $x   \in   \{0,1\}^*$ we
have ${\bf M} (x)/ \mu (x)   \geq   2^{{-} K( \mu )}$
by Theorem 4.5.1 and Equation 4.11 in \cite{LiVibook}.
\end{proof}
\begin{example}
\rm
Suppose we are given an infinite decimal sequence $\omega$.
The even positions contain the subsequent digits of
$\pi = 3.1415 \ldots $, and the odd positions
contain uniformly distributed,
independently drawn random decimal digits. Then,
${\bf M} (a| \omega_{1:2i} )   \rightarrow   1/10$
for $a = 0, 1,  \ldots , 9$, while
${\bf M} (a| \omega_{1:2i+1} )  \rightarrow   1$
if $a$ is the $i$th digit of $\pi$, and to 0 otherwise.
\end{example}
The universal distribution combines a  weighted version of the predictions
of all enumerable semimeasures, including the prediction
of the semimeasure with the shortest program. It is not
a priori clear that the shortest program dominates in all
cases---and as we shall see it does not. However, we
show that in the overwhelming majority of cases---the typical cases---the
shortest
program dominates sufficiently to 
use shortest programs for prediction. 

Taking the negative logarithm on both sides of (\ref{eq3}),
we want to determine $y$ with $l(y) = n$ that minimizes
\[
- \log \mu (y|x)= - \log {\mu (xy)} + \log {\mu (x)} .
\]
This $y$ is the most probable extrapolation of $x$. 
\begin{definition}\label{def.compl.monotoneKm}
\rm
Let $U$ be the reference monotone machine.
The complexity ${\it Km}$, called
{\it monotone complexity},\index{complexity!monotone|bold}
is defined as\index{${\it Km}$: monotone complexity}
$$
{\it Km} (x) = \min \{ l(p) : U(p) = x \omega, \omega \in \{0,1\}^{\infty} \}  .
$$
\end{definition}
We omit the Invariance Theorem for 
${\it Km}$ complexity, stated and proven
completely analogous to the Theorems with respect to the $C$ and $K$ varieties.
\begin{theorem}\label{theo.predcompr}
Let $\mu$ be a recursive semimeasure, and 
 let $\omega$ be a $\mu$-random infinite binary sequence and $xy$
be a finite prefix of $\omega$.
For $l(x)$ grows unboundedly and $l(y)$ fixed,
\[
\lim_{l(x) \rightarrow \infty}  - \log \mu (y|x) \eqa
Km(xy)- Km(x) < \infty, \]
where $Km(xy)$
and $Km(x)$ grow unboundedly.
\end{theorem}
\begin{proof}
By definition, $- \log {\bf M}(x) \leq {\it Km}(x)$ since
the left-hand side of the inequality weighs
the probability of {\em all} programs that
produce $x$ while the right-hand side weighs the 
probability of the shortest program only.
In the discrete case we have 
the Coding Theorem~\ref{PR2}:
$K (x) \eqa - \log {\bf m} (x)$.
L.A. Levin, \cite{Le73} erroneously
conjectured that also  $ {\it Km}(x) \eqa - \log {\bf M}(x)$.
But P. G\'acs  \cite{Ga83} showed that they are different, although
the differences 
must in some sense be very small:
\begin{claim}\label{lem.gacs1}
\begin{eqnarray}
&& - \log {\bf M} (x)   \leq  {\it Km} (x)   \lea  - \log  {\bf M} (x) 
+ {\it Km}(l(x));
\label{E.3.1} \\
&& \sup_{{x}   \in  \{0,1\}^*} |{\it KM} (x) - {\it Km}(x)| = \infty .
\nonumber
\end{eqnarray}
\end{claim}
However, for {\it a priori} almost all infinite sequences
$x$, the difference between ${\it Km} (\cdot)$ 
and $- \log {\bf M} (\cdot)$ is bounded
by a constant
\cite{Ga83}:
\begin{claim}\label{lem.gacs2}
(i) For random strings $x \in \{0,1\}^*$ we have 
${\it Km}(x)+ \log {\bf M}(x) \eqa 0$.

(ii)
There exists a function $f(n)$ which goes
to infinity with $n \rightarrow \infty$ such that
${\it Km}(x) +  \log {\bf M}(x) \geq f(l(x))$, for infinitely many $x$.
If $x$ is a finite binary string, then we can choose $f
(n)$
as the inverse of some version
of Ackermann's function
\end{claim}

 Let $\omega$ be a $\mu$-random infinite binary sequence and $xy$
be a finite prefix of $\omega$.
For $l(x)$ grows unboundedly with $l(y)$ fixed,
we have by Theorem~\ref{PR10}:
\begin{equation}\label{eq.muM}
\lim_{l(x) \rightarrow \infty}  \log \mu (y|x)  - \log {\bf M} (y|x) =0.
\end{equation}
Therefore, if $x$ and $y$ satisfy above conditions,
then maximizing $\mu (y|x)$ over $y$ means minimizing
 $- \log {\bf M} (y|x)$.
It is shown in Claim~\ref{lem.gacs2}
that $- \log {\bf M}(x)$ is slightly smaller than
$Km(x)$, the length of the shortest program for $x$ 
on the reference universal monotonic machine. For binary programs
this difference is very small, Claim~\ref{lem.gacs1}, but can be unbounded in
the length of $x$. 

Together this shows the following. Given $xy$ that is a prefix
of a (possibly not $\mu$-random) $\omega$,
optimal prediction of fixed length extrapolation $y$ from an
unboundedly growing prefix $x$ of $\omega$ need
not necessarily be achieved by the shortest programs for $xy$ and $x$
minimizing $Km(xy)-Km(x)$,
but is achieved by considering the weighted version of all
programs for $xy$ and $x$ which is represented by
$$- \log {\bf M} (xy)+ \log {\bf M} (x)
 = (Km(xy) - g(xy))-(Km(x)-g(x)).$$
Here $g(x)$ is a function which can rise to
in between the inverse of the Ackermann
function and $Km(l(x)) \leq \log \log x$---but only 
in case $x$ is not $\mu$-random.

Therefore, for certain $x$ and $y$ which are {\em not} $\mu$-random,
optimization using the minimum length programs may result
in incorrect predictions. 
For $\mu$-random $x$ we have that $- \log {\bf M} (x)$
and $Km(x)$ coincide up to an additional constant independent of $x$,
that is, $g(xy)=g(x) \eqa 0$,
Claim~\ref{lem.gacs2}.
Hence, together with Equation~\ref{eq.muM},
the theorem is proven.
\end{proof}

By its definition $Km$ is monotone in the sense
that always $Km(xy) - Km(x) \geq 0$. The closer this difference is to
zero, the better the shortest effective monotone program for $x$ is
also a shortest effective monotone program for $xy$ and hence
predicts $y$ given $x$.
Therefore, for  all large enough $\mu$-random $x$,
predicting by determining $y$
which minimizes the difference of the minimum program lengths of $xy$ and $x$
gives a good prediction. Here $y$ should be preferably large enough
to eliminate the influence of the $O(1)$ term.
\begin{corollary}[Prediction by Data Compression]
Assume the conditions of Theorem~\ref{theo.predcompr}. With $\mu$-probability
going to one as $l(x)$ grows unboundedly, a fixed-length $y$ extrapolation
from $x$ maximizes $\mu (y|x)$ iff $y$ can be maximally compressed
with respect to $x$ in the sense that it minimizes $Km(xy) - Km(x)$.
That is, $y$ is the string that minimizes the length
difference between the shortest program that outputs $xy \dots$
and the shortest program that outputs $x \ldots$.
\end{corollary}

\section{Conclusion}
The analysis of both hypothesis identification by ideal MDL and 
prediction shows that maximally compressed descriptions give
good results on the data samples which are random with respect
to probabilistic hypotheses. These data samples form the
overwhelming majority and occur with probability going
to one when the length of the data sample grows unboundedly.
\section*{Acknowledgement} Ray Solomonoff suggested to also analyze
the case of prediction. Jorma Rissanen, Chris Wallace, David Dowe,
Peter Gr\"unwald, Kevin Korb, and the anonymous referees
 gave valuable comments. Part of this work was
performed during a stay of the first author
 at Monash University, Melbourne, Australia.

\appendix
\section{Appendix: Kolmogorov Complexity}
\label{app.A}
\label{sect.kc}
The Kolmogorov complexity \cite{Ko65} of a finite object $x$
is the length of the
shortest effective binary description of $x$.
We give some definitions to establish notation.
For more details see \cite{ZvLe70,LiVibook}.
Let $x,y,z \in {\cal N}$, where
${\cal N}$ denotes the natural
numbers and we identify
${\cal N}$ and $\{0,1\}^*$ according to the
correspondence 
\[(0, \epsilon ), (1,0), (2,1), (3,00), (4,01), \ldots \]
Here $\epsilon$ denotes the {\em empty word} `' with no letters.
The {\em length} $l(x)$ of $x$ is the number of bits
in the binary string $x$. For example,
$l(010)=3$ and $l(\epsilon)=0$. 

The emphasis is on binary sequences only for convenience;
observations in any alphabet can be so encoded in a way
that is `theory neutral'.

A binary string $x$
is a {\em proper prefix} of a binary string $y$
if we can write $x=yz$ for $z \neq \epsilon$.
 A set $\{x,y, \ldots \} \subseteq \{0,1\}^*$
is {\em prefix-free} if for any pair of distinct
elements in the set neither is a proper prefix of the other.
A prefix-free set is also called a {\em prefix code}.
Each binary string $x=x_1 x_2 \ldots x_n$ has a
special type of prefix code, called a
{\em self-delimiting code},
\[ \bar x =x_1x_1x_2x_2 \ldots x_n \neg x_n ,\]
where
$\neg x_n=0$ if $x_n=1$ and $\neg x_n=1$ otherwise. This code
is self-delimiting because we can determine where the
code word $\bar x$ ends by reading it from left to right without
backing up. Using this code we define
the standard self-delimiting code for $x$ to be
$x'=\overline{l(x)}x$. It is easy to check that
$l(\bar x ) = 2 n$ and $l(x')=n+2 \log n$. 

Let $T_1 ,T_2 , \ldots$ be a standard enumeration
of all Turing machines, and let $\phi_1 , \phi_2 , \ldots$
be the enumeration of corresponding functions
which are computed by the respective Turing machines.
That is, $T_i$ computes $\phi_i$.
These functions are the {\em partial recursive} functions
or {\em computable} functions. The Kolmogorov complexity 
$C(x)$ of $x$ is the length of the shortest binary program
from which $x$ is computed. Formally, we define this as follows.

\begin{definition}\label{def.KolmC}
{\rm
The {\em Kolmogorov complexity} of $x$ given $y$ (for 
free on a special input tape) is
\[C(x|y) = \min_{p,i}\{l(i'p): \phi_i (p,y )=x , p \in \{0,1\}^*, i
\in {\cal N} \}. \]
Define $C(x)=C(x|\epsilon)$.}
\end{definition}

Though defined in terms of a
particular machine model, the Kolmogorov complexity
is machine-independent up to an additive
constant
 and acquires an asymptotically universal and absolute character
through Church's thesis, from the ability of universal machines to
simulate one another and execute any effective process.
  The Kolmogorov complexity of an object can be viewed as an absolute
and objective quantification of the amount of information in it.
   This leads to a theory of {\em absolute} information {\em contents}
of {\em individual} objects in contrast to classic information theory
which deals with {\em average} information {\em to communicate}
objects produced by a {\em random source} \cite{LiVibook}.

For technical reasons we also need a variant of complexity,
so-called prefix complexity, which associated with Turing machines
for which the set of programs resulting in a halting computation
is prefix free. We can realize this by equiping the Turing
machine with a one-way input tape, a separate work tape,
and a one-way output tape. Such Turing 
machines are called prefix machines
since the halting programs for anyone of them form a prefix free set.
Taking the universal prefix machine $U$ we can define
the prefix complexity analogously with the plain Kolmogorov complexity.
If $x^*$ is the first shortest program for $x$ then the set
$\{x^* : U(x^*)=x, x \in \{0,1\}^*\}$ is a {\em prefix code}.
That is, each $x^*$ is a code word for some $x$, and if $x^*$
and $y^*$ are code words for $x$ and $y$ with $x \neq y$ then $x^*$ is not
a prefix of $x$.

Let $\langle \cdot \rangle$ be a standard invertible
effective one-one encoding from ${\cal N} \times {\cal N}$
to prefix-free recursive subset of ${\cal N}$.
For example, we can set $\langle x,y \rangle = x'y'$.
We insist on prefix-freeness and
recursiveness because we want a universal Turing
machine to be able to read an image under $\langle \cdot \rangle$ 
from left to right and
determine where it ends.

\begin{definition}\label{def.KolmK}
{\rm
The {\em prefix Kolmogorov complexity} of $x$ given $y$ (for 
free) is
\[K(x|y) = \min_{p,i}\{l(\langle p,i\rangle): \phi_i (\langle p,y \rangle )=x , p \in \{0,1\}^*, i
\in {\cal N} \}. \]
Define $K(x)=K(x|\epsilon)$.}
\end{definition}

The nice thing about $K(x)$ is that we can interpret $2^{-K(x)}$
as a probability distribution. Namely, $K(x)$ is the length of
a shortest prefix-free program for $x$. By the fundamental
Kraft's inequality, see for example \cite{CT91,LiVibook}, we know that
if $l_1 , l_2 , \ldots$ are the code-word lengths of a  prefix code,
then $\sum_x 2^{-l_x} \leq 1$. This leads to the notion
of universal distribution---a rigorous form of Occam's razor--below.

\section{Appendix: Universal Distribution}
\label{app.B}
A Turing machine $T$ computes a function on the natural numbers.
However, we can also consider the computation
of real valued functions. For this purpose we consider
both the argument of $\phi$ and the value of $\phi$
as a pair of natural numbers according to the standard
pairing function $\langle \cdot \rangle$. We define
a function from ${\cal N}$ to the reals ${\cal R}$
by a Turing machine $T$ computing
a function $\phi$ as follows. Interprete
the computation $\phi(\langle x,t \rangle ) = \langle p,q \rangle$
to mean that the quotient $p/q$ is
the rational valued $t$th approxmation of $f(x)$.

\begin{definition}\label{def.enum.funct}
{\rm
A function $f: {\cal N} \rightarrow {\cal R}$ is
{\em enumerable} if there is a Turing machine $T$ computing a
total function $\phi$ 
such that $\phi (x,t+1) \geq \phi (x,t)$ and
$\lim_{t \rightarrow \infty} \phi (x,t)=f(x)$. This means
that $f$ can be computably approximated from below.
If $f$ can also be computably approximated from above
then we call $f$ {\em recursive}.}
\end{definition}

A function $P: {\cal N} \rightarrow [0,1]$ is
a {\em probability distribution} if
$\sum_{x \in {\cal N}} P(x) \leq 1$. (The inequality
is a technical convenience. We can consider
the surplus probability to be concentrated on the
undefined element $u \not\in {\cal N}$).

Consider the family ${\cal EP}$ of
{\it enumerable} probability distributions on the
sample space ${\cal N}$ (equivalently, $\{0,1\}^*$).
It is known, \cite{LiVibook}, that ${\cal EP}$
contains an element $\hbox{\bf m}$ that
multiplicatively dominates all elements of ${\cal EP}$. That is,
for each $P \in {\cal EP}$ there is a constant $c$ such
that $c \: \hbox{\bf m} (x) > P(x)$ for all $x \in {\cal N}$.
We call ${\bf m}$ a {\em universal distribution}.

The family ${\cal EP}$ contains all distributions
with computable parameters
which have a name, or in which we could conceivably
be interested, or which have ever been considered.
The dominating property means that $\hbox{\bf m}$
assigns at least as much probability to each object
as any other distribution in the family ${\cal EP}$
does. In this sense it is a universal {\em a priori}
by accounting for maximal ignorance. It turns out that
if the true {\em a priori} distribution in Bayes's rule
is recursive, then using the single distribution
$\hbox{\bf m}$, or its continuous analogue the measure $\hbox{\bf M}$
on the sample space $\{0,1\}^{\infty}$ (Section~\ref{sect.predict}),  
is provably as good
as using the true {\em a priori} distribution.

We also know, \cite{Le74,Ga74,Ch75}, that
\begin{lemma}\label{PR2}
\begin{equation}\label{eq.m}
 - \log \hbox{\bf m} (x)=K(x) \pm  O( 1). 
\end{equation}
\end{lemma}
That means that $\hbox{\bf m}$ assigns high probability to simple 
objects
and low probability to complex or random objects.
For example, for $x=00 \ldots 0$ ($n$ 0's) we have
$K(x) \eqa K(n) \lea \log n + 2 \log \log n $ since the program
\[ \mbox{\tt print } n \mbox{\tt \_times a ``0''} \]
prints $x$. (The additional $2 \log \log n$ term
is the penalty term for a self-delimiting encoding.)
Then, $1/ (n \log^2 n ) = O( \hbox{\bf m}(x))$.
But if we flip a coin to obtain a string $y$ of $n$ bits,
then with overwhelming probability $K(y) \gea n $
(because $y$ does not contain effective regularities
which allow compression),
and hence $\hbox{\bf m}(y) = O( 1/2^n)$.
\section{Appendix: Randomness Tests}\label{sect.rt}
\label{app.C}
One can consider those objects as nonrandom
in which one can find sufficiently many
regularities.
In other words, we
would like to identify 
``incompressibility''  with
``randomness.''
This is proper if
the sequences that are incompressible can be shown to possess
the various properties of randomness (stochasticity)
known from the theory of probability. 
That this is possible is the substance of
the celebrated theory developed by the Swedish
mathematician
Per Martin-L\"of \cite{ML66}. This theory was further elaborated
in \cite{ZvLe70,Sch71,KU87} and later papers.

There are many properties known which 
probability theory attributes to random
objects. To give an example,
consider sequences of $n$ tosses with a fair coin.
Each sequence of $n$ zeros and ones is
equiprobable as an outcome: its
probability is $2^{-n}$.
If such a sequence is to be random
in the sense of a proposed new definition, then
the number of ones in $x$ should be near to $n/2$,
the number of occurrences of blocks ``00'' 
should be close to $n/4$, and so on.

It is not
difficult to show that each such single property separately holds
for all incompressible binary strings.
But we want to demonstrate
that incompressibility implies all
conceivable effectively testable properties of randomness
(both the known ones and the as yet unknown
ones). This way, the various theorems 
in probability theory about random sequences
carry over automatically to
incompressible sequences.
\begin{comment}
In the case of finite strings we cannot hope
to distinguish sharply between random and
nonrandom strings. For instance, considering the set of 
binary strings of
a fixed length, it
would not be natural to
to fix an $m$ and call a string with $m$ zeros random
and a string with $m+1$ zeros nonrandom.
%
\end{comment}
Let us borrow some ideas from statistics.
We are given a certain sample space $S$ with an associated distribution $P$.
Given an element $x$ of the sample space,
we want to test the hypothesis ``$x$ is a typical outcome.''
Practically speaking, the property of being typical is
the property of belonging to any reasonable majority.
In choosing an object at random, we have confidence that
this object will fall precisely in the intersection
of all such majorities. The latter condition we identify
with $x$ being random.

To ascertain whether a given element of the sample space
belongs to a particular reasonable majority we introduce the notion of a test.
\index{test!in statistics|bold}
Generally, a test is given by a prescription which, for
every level of significance $\epsilon$, tells us for what
elements $x$ of $S$ the hypothesis ``$x$ belongs to majority $M$ in $S$''
should be rejected, where $\epsilon = 1 - P(M)$. 
Taking $\epsilon = 2^{{-} m}$, $m = 1, 2, \ldots $,
this amounts to saying that we have a description
of the set $V   \subseteq   {\cal N} \times S$ of nested %
\it critical regions %
\index{test!critical region of|bold}
\rm
\begin{eqnarray*}
 V_m & = &  \{  x: (m, x)   \in   V  \} \\
 V_m  &  \supseteq  & V_{{m} + 1} ,   \ \ \ \  m = 1, 2,  \ldots .
\end{eqnarray*}
The condition that $V_m$ be a critical region on the
{\it significance level}\index{test!significance level of}
$\epsilon  = 2^{-m}$ amounts to requiring, for all $n$
$$
\sum_x    \{  P(x): l(x) = n, x   \in   V_m   \}    \leq \epsilon .
$$
The complement of a critical region $V_m$ is called the
$(1- \epsilon )$ {\it confidence interval}.
\index{test!confidence interval of}
If $x \in V_m$, then
the hypothesis ``$x$ belongs to majority $M$,'' and therefore
the stronger hypothesis ``$x$ is random,'' is rejected with
\index{test!testing for randomness}
significance level $\epsilon $.
We can say that $x$ fails the test at the level of 
critical region $V_m$. 

\begin{example}\label{example.MLtest}
\rm
A string $x_1 x_2  \ldots x_n$ 
with many initial zeros is not very
random. We can test this aspect as follows.
The special test $V$ has
critical regions $V_1 , V_2 , \ldots $. 
Consider $x = 0.x_1 x_2  \ldots x_n$ as a rational number,
and each critical region as a half-open interval
$V_m = [0, 2^{-m}  )$ in $[0, 1)$, $m = 1, 2, \ldots $.
Then the subsequent
critical regions test the hypothesis ``$x$ is random'' by considering
the subsequent digits in the binary expansion of $x$.
We reject the hypothesis on the 
significance level $\epsilon = 2^{-m}$
provided $x_1 = x_2 = \cdots = x_m = 0$,
\end{example}

\begin{example}\label{example.anothertest}
\rm
Another test for randomness of finite binary
strings rejects when the relative frequency
of ones differs too much from $1/2$.
This particular test can be implemented by
rejecting the hypothesis of randomness of
$x=x_1 x_2 \ldots x_n$ at level $\epsilon  = 2^{-m}$ provided
$|2 f_n - n|   >   g(n, m)$,
where $f_n = \sum_{i=1}^n x_i$,
and $g(n, m)$ is the least number determined by the requirement
that the number of binary strings $x$ 
of length $n$ for which this inequality holds
is at most $2^{n-m}$.
\end{example}

In practice, statistical tests are %
\it effective %
\rm prescriptions
such that we can compute, at each level of significance,
for what strings the associated hypothesis should be rejected.
It would be hard to imagine what use
it would be in statistics to have tests
that are not effective in the sense of computability
theory.

\begin{definition}
\rm
\label{test}
Let $P$ be a recursive
probability distribution
on the sample space ${\cal N}$.
A total\index{test|bold}
function $\delta : {\cal N}   \rightarrow   {\cal N}$ is
a $P$-{\em test} (Martin-L\"of 
test\index{test!Martin-L\"of|see{test}} 
for randomness)\index{test!$P$-|see{test}} if:
\begin{enumerate}
\item
$\delta$ is enumerable
(the set
$V =   \{  (m, x): \delta (x)   \geq   m  \}  $ %
\rm is recursively
enumerable); and
\item
$\sum   \{  P(x): \delta (x)   \geq   m ,$
$l(x) =$
$n   \}    \leq $
$2^{-m}$, for all $n$.
\end{enumerate}
\end{definition}

The critical regions associated with 
the common statistical tests
are present in the form
of the sequence 
$V_1   \supseteq   V_2   \supseteq    \cdots $, where
$V_m =   \{  x: \delta (x)   \geq   m  \}  $%
\rm , for $m   \geq   1$.
Nesting is assured since $\delta (x)   \geq   m+1$ implies
$\delta (x)   \geq   m$. Each set $V_m$ is recursively enumerable
because of Item 1.

A particularly important case is $P$ is
the uniform distribution\index{distribution!uniform}, defined by
$L(x) = 2^{-2l(x)}$. The restriction of $L$ to strings of length
$n$ is defined by $L_n (x)=2^{-n}$ for $l(x)=n$ and 0 otherwise.
(By definition, $L_n (x)=L(x|l(x)=n)$.)
Then, Item 2 can be rewritten as $\sum_{x \in V_m} L_n (x) \leq 2^{-m}$
which is the same as
$$
d(  \{  x: l(x) = n, \: x   \in   V_m    \}  )  \leq 2^{n-m}.
$$
In this case we often speak simply of a %
\it test\index{test}%
\rm ,
with the uniform distribution $L$ understood.

\begin{comment}
In statistical
tests 
membership of $(m,x)$
in $V$ can usually be determined in polynomial time in $l(m) + l(x)$.
\end{comment}

\begin{example}\label{example.test.oddones}
\rm
The previous test examples can be rephrased 
in terms of Martin-L\"of tests.
Let us try a more subtle example. A real number such that
all bits in odd positions in its binary representation are 1's
is not random with respect to the uniform distribution.
To show this we need a test which detects
sequences of the form $ x =  1 x_2 1 x_4 1 x_6 1 x_8   \ldots $.
Define a test $\delta$ by
$$
\delta (x)  = \max   \{   i:  x_1 = x_3 =  \cdots  = x_{2i-1} = 1  \}  ,
$$
and $\delta (x) = 0$ if $x_1 = 0$.
For example: $\delta (01111) = 0$; $\delta (10011) = 1$; $\delta (11011) = 1$;
$\delta (10100) = 2$; $\delta (11111) = 3$.
To show that $\delta$ is a test we have to show that $\delta$ 
satisfies the definition of a test. Clearly, $\delta$ is
enumerable (even recursive). 
If $\delta (x)   \geq   m$ where $l(x) = n   \geq   2m$,
then there are $2^{m-1}$ possibilities for the $(2m - 1)$-length
prefix of $x$, and $2^{n-(2m-1)}$ possibilities for the remainder
of $x$. Therefore,
$d  \{  x: \delta (x)   \geq   m,  \: l(x) = n  \}   \leq 2^{n-m}$.
\end{example}

\begin{definition}\label{def.martinloeftest}
\rm
A universal Martin-L\"of test for randomness with respect
to distribution $P$, a %
\it universal P-test %
\index{test!universal|bold}
\rm for short,
is a test $\delta_0 (\cdot |P)$ %
\rm  such that for
each $P$-test $\delta$, there is a constant $c$,
such that for all $x$, we have $\delta_0 (x|P)   \geq   \delta (x) - c$.
\end{definition}

\begin{comment}
We say that $\delta_0(\cdot |P)$ (additively) 
majorizes $\delta$%
\rm . Intuitively, $\delta_0(\cdot |P)$
constitutes a test for randomness
which incorporates
all particular tests $\delta$
in a single test.
No test for randomness $\delta$ other than $\delta_0(\cdot |P)$ 
can discover more than a constant
amount more deficiency of 
randomness\index{randomness deficiency} in any string $x$.
In terms of critical regions\index{test!critical region of}, 
a universal test is a test such that
if a binary sequence is random with respect to that test,
then it is random with respect to any conceivable test,
neglecting a change in significance level. Namely,
with $\delta_0(\cdot |P)$ a universal $P$-test,
let $U  =   \{  (m, x): \delta_0 (x|P)   \geq   m  \}  $, and, for any
test $\delta$, let $V  =   \{  (m, x): \delta (x)   \geq   m  \}  $.
Then, defining the associated critical zones as before,
we find
$$
V_{{m} + c}   \subseteq   U_m ,  \;   m = 1, 2,  \ldots  ,
$$
where $c$ is a constant (dependent only on $U$ and $V$).
\end{comment}
It is a major result that there exists a universal $P$-test.
The proof goes by first showing that the set of all
tests is enumerable.

\begin{lemma}\label{lemma.Ptest}
We can effectively enumerate all $P$-tests.
\end{lemma}
\begin{proof}
We start with the standard enumeration
$\phi_1 , \phi_2 , \ldots $ of partial recursive
functions from ${\cal N}$ into ${\cal N} \times {\cal N}$,
and turn this into an enumeration $\delta_1 , \delta_2 , \ldots$
of all and only $P$-tests.
The list $\phi_1 , \phi_2 , \ldots $ 
enumerates all and only recursively enumerable
sets of pairs of integers as
$\{ \phi_i (x): x \geq 1 \}$ for $i=1,2, \ldots$
In particular, for any $P$-test $\delta$, the set
$  \{  (m, x): \delta (x)   \geq   m  \}  $ occurs in this list.
The %
\it only %
\rm thing we have to do is to eliminate those $\phi_i$
of which the range does not correspond to a $P$-test.

First, we effectively modify
each $\phi$ (we drop the subscript for convenience)
to a function $\psi $ such that range $\phi$ equals
range $\psi$, and $\psi$ has the special
property that if $\psi (n)$ is defined,
then $\psi (1), \psi (2), \ldots , \psi (n-1)$ are
also defined. This can be done by dovetailing
the computations of $\phi$ on the different arguments:
in the first phase do one step of the computation of $\phi (1)$,
in the second phase do the second step of the computation of $\phi (1)$
and the first step of the computation of $\phi (2)$.
In general, in the $n$th phase we execute the $n_1$th step
of the computation of $\phi (n_2 )$, for all $n_1 , n_2$
satisfying $n_1 + n_2 = n$. We now define $\psi$ as follows.
If the first computation that halts is that of $\phi (i)$, then set
$\psi (1) := \phi (i)$. If the second computation that
halts is that of $\phi (j)$, then set $\psi (2) := \phi (j)$, and so on.

Secondly, use each $\psi$
to construct a test $\delta$ by
approximation from below.
In the algorithm, at each stage of the computation the local
variable array 
$\delta (1: \infty )$ contains the current 
approximation
to the list of function values $\delta (1), \delta (2), \ldots$.
This is doable because the nonzero part of the approximation
is always finite.

\begin{description}
\item[Step 1]
Initialize $\delta$ by setting $\delta (x) := 0$ for all $x$.
$\{$If the range of $\psi$ is empty, then
this assignment will not be changed in 
the remainder of the procedure.
That is, $\delta$ stays identically zero and
it is trivially a test.$\}$ Initialize $i := 0$.
\item[Step 2]
Set $i := i + 1$; compute $\psi (i)$ and let its value be $(x, m)$.
\item[Step 3]
If $\delta (x)   \geq   m$ then go to Step 2.
else set $\delta (x) := m$.
\item[Step 4]
If $\sum    \{  P(y): \delta (y) \geq k, l(y) = l(x)  \}  >2^{{-} k}$
for some $k$, $k =  1,  \ldots , m$
$\{$since $P$ is a recursive function
we can effectively test whether the new value of $\delta (x)$ violates
Definition~\ref{def.martinloeftest}$\}$ 
then set $\delta (x) := 0$ and terminate 
$\{$the computation of $\delta$ is finished$\}$
else go to Step 2.
\end{description}

(With $P$ the uniform distribution\index{distribution!uniform},
for $i = 1$ the conditional in Step 4 simplifies
to $m    >   l(x )$.)
In case the range of $\psi$ is already a test, then the
algorithm never finishes but forever
approximates $\delta$ from below.
If $\psi$ diverges for some argument then the computation
goes on forever and does not change $\delta$ any more.
The resulting $\delta$ is an
enumerable test.
If the range of $\psi$ is
not a test, then at some point the conditional in Step 4 is
violated and the approximation of $\delta$ terminates.
The resulting $\delta$ is a test, even a recursive one.
Executing this procedure on all functions in the list
$\phi_1 , \phi_2 , \ldots $, we obtain
an effective enumeration $\delta_1 , \delta_2 , \ldots $
of all $P$-tests (and only $P$-tests). We are now in the position
to define a universal $P$-test.
\end{proof}

\begin{theorem}
\label{M0}
\index{test!universal}
Let $\delta_1 , \delta_2 , \ldots $ be an enumeration
of above $P$-tests. Then,
$\delta_0 (x|P) = \max   \{   \delta_y (x) - y : y   \geq   1  \}  $ is
a universal $P$-test.
\end{theorem}

\begin{proof}
Note first that $\delta_0(\cdot |P)$ is a total function on $\cal N$
because of Item 2 in Definition~\ref{def.martinloeftest}.

(1) The enumeration $\delta_1 , \delta_2 , \ldots $ in
Lemma~\ref{lemma.Ptest} yields an
enumeration of recursively enumerable sets:
$$
  \{  (m, x): \delta_1 (x)   \geq   m  \}  ,  \
  \{  (m, x): \delta_2 (x)   \geq   m  \}  ,  \ldots .
$$
Therefore,
$V=  \{  (m, x): \delta_0 (x|P)   \geq   m  \}  $
is recursively enumerable.

(2) Let us verify
that the critical regions are small enough: for each $n$,
\begin{eqnarray*}
\sum_{l(x)=n} \{  P(x): \delta_0 (x|P)   \geq   m \}  &   \leq &
\sum_{{y} = 1}^{\infty}  \ \sum_{l(x)=n} \{  P(x): \delta_y (x) \geq  m+y \}\\
& \leq & \sum_{{y} = 1}^{\infty}  2^{-m-y} = 2^{-m} .
\end{eqnarray*}

(3) By its definition, $\delta_0(\cdot |P)$ majorizes
each $\delta$ additively. Hence, it is universal.
\end{proof}

By definition of $\delta_0(\cdot |P)$ as a universal $P$-test,
any particular $P$-test $\delta$ can discover at most a constant
amount more regularity in a sequence $x$ than
does $\delta_0(\cdot |P)$, in the sense that
for each $\delta_y$ we have
$\delta_y (x)  \leq \delta_0 (x|P)) + y$ for all $x$.

For any two universal $P$-tests $\delta_0(\cdot |P)$
and ${\delta '}_0(\cdot |P)$, there is a constant $c   \geq   0$,
such that for all $x$, we have
$| \delta_0 (x|P) - {\delta '}_0 (x|P)|  \leq c$.

We started out with the objective to establish in what sense
incompressible strings may be called random.

\begin{theorem}
\label{M1}
The function
$f(x) = l(x) - C(x| l(x))-1$
is a universal $L$-test with $L$ the uniform distribution.
\index{test!universal for uniform distribution}
\end{theorem}

\begin{proof}
(1)
We first show that $f(x)$ is a test with
respect to the uniform distribution.
The set $  \{  (m, x): f(x)   \geq   m  \}  $
is recursively enumerable since $C()$ can be
approximated from above by a recursive process.

(2) We verify the condition on the critical regions.
Since the number of $x$'s with $C(x| l(x))  \leq l(x) - m-1$
cannot exceed the number of programs of length at most
$l(x) - m-1$, we have
$d(  \{  x: f(x)   \geq   m  \}  )  \leq 2^{l(x)-m} -1$.

(3) We show that for each test $\delta$,
there is a constant $c$, such that
$f(x)   \geq   \delta (x) - c$. The main idea is
to bound $C(x| l(x))$ by
exhibiting a description of $x$, given $l(x)$.
Fix $x$. Let the set $A$ be defined as
$$
A =   \{  z: \delta (z)   \geq   \delta (x), l(z) = l(x)  \}   .
$$
We have defined $A$ such that $x   \in   A$ and
$d(A)  \leq 2^{{l(x)} - \delta (x) }$.
Let $\delta = \delta_y$ in the
standard enumeration $\delta_1 , \delta_2 , \ldots $ of tests.
Given $y$, $l(x)$, and $\delta (x)$,  we can enumerate all elements of $A$.
Together with $x$'s index $j$ in enumeration order of $A$,
this suffices to find $x$.
We pad the standard binary representation of $j$
with nonsignificant zeros to a string $s = 00 \ldots 0j$ of
length $l(x) - \delta (x)$. This is possible since
$l(s)   \geq   l(d(A))$. The purpose of changing $j$ to $s$
is that now the number $\delta (x)$ can be deduced
from $l(s)$ and $l(x)$. In particular,
there is a Turing
machine which
computes $x$ from
input $\bar y  s$, when $l(x)$ is given for free.
Consequently, since $C()$ is the shortest effective description,
$C(x| l(x))  \leq $
$l(x) - \delta (x) + 2l(y) + 1$.
Since $y$ is a constant
depending only on $\delta$,
we can set
$c = 2l(y) + 2$.
\end{proof}

In Theorem~\ref{M0},
we have exhibited a universal $P$-test for randomness of a string $x$
of length $n$ with respect to an arbitrary recursive distribution $P$
over the sample set
$S= {\cal B}^n$ with ${\cal B} =  \{ 0, 1 \}$.

The universal $P$-test
measures how justified is the assumption
\index{outcome of experiment}
that $x$ is the outcome of an experiment with distribution $P$.
We now use ${\bf m}$ to investigate alternative characterizations
of random elements of the
sample set $S= {\cal B}^*$ (equivalently, $S={\cal N}$).

\begin{definition}\label{def.(3.10)}
\rm
Let $P$ be a recursive probability distribution on ${\cal N}$.
A {\em sum $P$-test}\index{test!sum} is a nonnegative
enumerable function $\delta$ satisfying
\begin{equation}
\sum_x  P(x) 2^{{\delta} (x)}  \leq  1.
\label{(3.1)}
\end{equation}
A %
{\em universal sum $P$-test} \index{test!universal sum|bold}
is a test that additively dominates each sum $P$-test.
\end{definition}

The sum tests of Definition~\ref{def.(3.10)}
are slightly stronger than the tests according to
Martin-L\"of's original
Definition~\ref{test}.

\begin{lemma}\label{lem.spt}
Each sum $P$-test is a $P$-test. If $\delta (x)$
is a $P$-test, then there is a constant $c$
such that $\delta'(x)=\delta (x) - 2 \log \delta (x) -c$
is a sum $P$-test.
\end{lemma}

\begin{proof}
It follows immediately from the new
definition that for all $n$
\begin{equation}
\sum  \{ P(x): \delta (x) > k, l(x)=n \}   \leq  2^{-k} .
\label{(3.2)}
\end{equation}
Namely, if (\ref{(3.2)}) is false, then
we contradict (\ref{(3.1)}) by
$$
\sum_{{x}  \in  {\cal N}} P(x) 2^{{\delta} (x)} >
\sum_{{l(x)} = n} P(x) 2^k  \geq  1 .
$$

Conversely, if $\delta (x)$ satisfies (\ref{(3.2)}) for all $n$,
then for some constant $c$, the
function $\delta (x) - 2 \log \delta (x) - c$
satisfies (\ref{(3.1)}).
\end{proof}

This shows that the sum test
is not much stronger than the original test. One advantage
of (\ref{(3.1)}) is that it is just one inequality,
instead of infinitely many, one for each $n$.
We give an exact expression for a universal
sum $P$-test in terms of complexity.

\begin{theorem}\label{PR3}
Let $P$ be a recursive probability distribution.
The function $$\kappa_0 (x| P ) =  \log ( {\bf m} (x)/ P (x)) $$
is a universal sum $P$-test.
\index{test!universal sum}
\end{theorem}

\begin{proof}
Since ${\bf m}$ is enumerable, and $P$
is recursive, $\kappa_0 (x| P )$ is enumerable.
We first show that $\kappa_0 (x| P )$ is a sum $P$-test:
$$
\sum_x  P(x) 2^{\kappa_0(x|P)} =
\sum_x  {\bf m} (x)  \leq  1.
$$
It is only left to show that $\kappa_0 (x| P )$
additively dominates all sum $P$-tests. For each sum
$P$-test $\delta$, the function $P(x) 2^{{\delta} (x)}$
is a semimeasure that is enumerable.
It has been shown, Section~\ref{sect.kc},
that there is a positive constant $c$ such that
$c \cdot {\bf m} (x)  \geq  P(x) 2^{{\delta} (x)}$.
Hence, there is another constant $c$ such
that $c \cdot \kappa_0 (x|P) \geq \delta (x)$,
for all $x$.
\end{proof}

\begin{example}\label{exam.condit.universaltest}
\rm
An important case is as follows.
If we consider a distribution $P$ restricted to a
domain $A \subset {\cal N}$, then the universal sum $P$-test
becomes $\log ({\bf m}(x|A)/P(x|A))$.
For example, if $L_n$ is the uniform distribution
on $A= \{0,1\}^n$, then the universal sum $L_n$-test
for $x \in A$ becomes
\[ \kappa_0(x|L_n)= \log ({\bf m} (x|A)/L_n (x)) \eqa n- K(x|n) . \]
Namely, $L_n(x) = 1/2^n$ and
$\log {\bf m}(x|A) = - K(x|A)$ by the Coding Theorem, Section~\ref{sect.kc}.
where we can describe $A$ by giving $n$.
\end{example}

\begin{example}
\rm
The Noiseless Coding
Theorem states
that the Shannon-Fano code, which codes
a source word $x$ straightforwardly
as a word of about
$ - \log P(x)$ bits, Section~\ref{sect.kc},
nearly achieves
the optimal expected code word length.
This code is uniform in the sense that
it does not use any characteristics of $x$
itself to associate a code word with
a source word $x$.
The code that codes
each source word $x$ as a code word of
length $K(x)$ also achieves the optimal
expected code word length. This code
is nonuniform in that it uses
characteristics of individual $x$'s to
obtain shorter code words. Any difference in
code word length between
these two encodings for a particular object
$x$ can only be due to exploitation
of the individual regularities in $x$.

Define the
\it randomness deficiency
\rm of a finite
object $x$ %
\it with respect %
\rm to $P$ as
\[ - \lfloor \log P(x) \rfloor - K(x)
\eqa - \log P(x) + \log {\bf m} (x)
\eqa \kappa_0 (x|P) ,\]
by the major theorems in Section~\ref{sect.rt}.
That is, the randomness deficiency
is the outcome of the universal sum
$P$-test
of Theorem~\ref{PR3}.
\end{example}

\begin{example}
\rm
Let us compare the randomness
deficiency\index{randomness deficiency}
as measured by $\kappa_0 (x| P)$ with that
measured by the
universal test\index{test!universal for uniform distribution} $\delta_0 (x)$,
for the uniform distribution.
That test consisted actually
of tests for a whole family $L_n$ of distributions,
where $L_n$ is the uniform distribution such that
each $L_n (x) = 2^{{-} n}$ for $l(x) = n$, and zero otherwise.
Rewrite $\delta_0 (x)$ as
$$
\delta_0 (x| L_n ) = n - C(x| n),
$$
for $l(x) = n$, and $\infty$ otherwise.
This is close to the expression for
$\kappa_0 (x| L_n  )$
obtained in Example~\ref{exam.condit.universaltest}.
>From the relations between $C$ and $K$ in \cite{LiVibook}
it follows that
\[ |\delta_0 (x|L_n) - \kappa_0(x|L_n)| \lea
2 \log C(x). \]

The formulation of the universal sum
test\index{test!universal sum}
in Theorem~\ref{PR3} can be interpreted as follows.
An element $x$ is random with respect
to distribution $P$, that is, $\kappa_0 (x| P) \eqa $,
if $P(x)$ is large enough, not in absolute value but
relative to ${\bf m} (x)$. If we did not have this relativization,
then we would not be able to distinguish between random
and nonrandom outcomes for the uniform distribution
$L_n (x)$ above.

Let us look at an example. Let $x=00 \ldots 0$ of
length $n$. Then,
$\kappa_0 (x| L_n ) \eqa n -  K(x|n) 
\eqa n $.
If we flip a coin $n$ times to generate $y$, then with
overwhelming probability $K(y|n) \geq n$ and
$\kappa_0 (y| L_n )  =  O( 1)$.
\end{example}
\begin{example}\label{fermi-dirac}
\rm
According to modern physics,
electrons, neutrons and protons satisfy
the Fermi-Dirac distribution.
We distribute $n$ particles among $k$
cells, for $n \leq k$,
such that
each cell is occupied by at most one particle; and
all distinguished arrangements satisfying this
have the same probability.

We can treat each arrangement as a binary string: an empty cell is a
zero and a cell with a particle is a one.
Since there are ${k \choose n }$ possible arrangements,
the probability for each arrangement $x$ to happen, under the
Fermi-Dirac distribution, is
${\it FD}_{n,k} (x)= {k \choose n}^{-1}$.
According to Theorem~\ref{PR3}:
\[\kappa_0 (x|{\it FD}_{n,k}) =
\log ( {\bf m}(x|k,n)/{\it FD}_{n,k}(x) ) 
\eqa -K(x|n,k)+\log {k \choose n}\]
is a universal sum test with respect to the
Fermi-Dirac distribution. It is easy to see that a
binary string $x$ of length $k$
with $n$ ones has complexity
$K(x|n,k) \leq \log {k \choose n}$,
and $K(x|n,k) \gea \log {k \choose n}$
for most such $x$.
Hence, a string $x$ with maximal $K(x|n,k)$
will pass this universal sum test.
Each individual such string possesses all effectively
testable properties of typical strings
under the Fermi-Dirac distribution. Hence, in
the limit for $n$ and $k$ growing unboundedly, we cannot effectively
distinguish one such a string from other such strings.
\end{example}
\begin{example}\label{exam.markov}
\rm
\it Markov's Inequality 
\rm says the following.
Let $P$ be any probability distribution, $f$ any nonnegative
function with $P$-expected value
${\bf E} = \sum_x P(x) f(x) < \infty$.
For ${\bf E}   \geq  0$ we have
$\sum  \{ P(x): f(x)/{\bf E} > k \}  < 1/k$.

Let $P$ be any probability distribution (not
necessarily recursive). The $P$-expected
value of ${\bf m} (x)/P(x)$ is
$$
\sum_x  P(x) \frac{{\bf m} (x)}{P (x)} \leq 1.
$$
Then, by Markov's Inequality
\begin{equation}
\sum_x \{ P(x): {\bf m} (x)  \leq  k P(x) \}   \geq  1  - {1 \over k }.
\label{(3.3)}
\end{equation}

Since ${\bf m}$ dominates all enumerable
semimeasures multiplicatively,
we have for all $x$,
\begin{equation}
P(x)  \leq  c_P  {\bf m} (x), \mbox{ and it can be shown  } c_P = 2^{K(P)} .
\label{(3.4)}
\end{equation}

Equations~(\ref{(3.3)}, \ref{(3.4)})
have the following consequences.

\begin{enumerate}
\item
If $x$ is a random sample from a simple recursive
distribution $P$, where ``simple'' means that $K(P)$
is small, then ${\bf m}$ is a good estimate
for $P$. For instance, if $x$ is randomly
drawn from distribution $P$, then the probability
that
\[c_P^{-1} {\bf m} (x) \leq P(x) \leq c_P {\bf m} (x)\]
is at least $1-1/c_P$.
\item
If we know or believe that $x$ is random with respect
to $P$, and we know $P(x)$, then we can use $P(x)$
as an estimate of ${\bf m} (x)$.
\end{enumerate}

In both cases the degree of approximation
depends on the index of $P$, and
the randomness of $x$ with respect to $P$,
as measured by the randomness deficiency
$\kappa_0 (x| P) = \log ( {\bf m} (x)/P(x))$.
For example, the
uniform discrete distribution on ${\cal B}^*$ can be defined by
\index{distribution!uniform discrete} $L (x) =  2^{-2l(x)}$.
Then, for each $n$ we have $L_n(x)=L(x|l(x)=n)$. To describe $L$
takes $O(1)$ bits, and therefore
$$
\kappa_0 (x| L ) \eqa l(x) - K(x).
$$
The randomness deficiency $\kappa_0 (x| L) \eqa 0$ iff
$K(x) \gea l(x)$, that is, iff $x$ is random.

The nonrecursive ``distribution'' ${\bf m} (x) = 2^{-K(x)}$
has the remarkable property that the test
$\kappa_0 (x| {\bf m} ) \eqa 0$ for all $x$: the test
shows all outcomes $x$ random with respect to it. We
can interpret (\ref{(3.3)}, \ref{(3.4)})
as saying that if the real distribution
is $P$, then $P (x)$ and ${\bf m} (x)$ are close to each other
with large $P$-probability. Therefore, if $x$ comes from some unknown
recursive distribution $P$, then we can use ${\bf m} (x)$ as an
estimate for $P (x)$. In other words, ${\bf m} (x)$ can be viewed
as the universal
\it a priori %
\rm probability'\index{probability!universal prior} of $x$.

The universal sum $P$-test\index{test!universal} $\kappa_0 (x|  P )$
can be interpreted in the framework of hypothesis testing as
the likelihood ratio between hypothesis $P$ and the fixed
alternative hypothesis ${\bf m}$. In ordinary statistical
hypothesis testing, some properties of an unknown distribution $P$
are taken for granted, and the role of the universal test can
probably be reduced to some tests that are used in statistical
practice.
\end{example}

\bibliographystyle{amsplain}

\end{document}